\documentclass[11pt]{article}

\usepackage[]{acl}

\usepackage{latexsym}
\usepackage{CJKutf8}

\usepackage{microtype}
\usepackage{natbib}
\usepackage{inconsolata}

\usepackage{graphicx}
\usepackage{amsmath}
\usepackage{amsfonts}
\usepackage{xspace}
\usepackage{lipsum,tabularx}

\usepackage{multicol}
\usepackage{multirow}
\usepackage{colortbl}
\usepackage{booktabs}
\usepackage{setspace}
\hypersetup{
  linkcolor={blue!70!black},
  citecolor={red!70!black},
  urlcolor={blue!70!black}
}
\def\Snospace~{\S{}}

\usepackage{algorithm}
\usepackage{algorithmicx}
\usepackage[noend]{algpseudocode}

\usepackage{balance}

\usepackage{bm}
\usepackage{fp}
\usepackage{siunitx}
\usepackage{amsthm}

\usepackage[labelfont=bf,font=small,skip=5pt]{caption}
\usepackage{subcaption}

\usepackage{comment}

\usepackage{fancyhdr}
\usepackage{framed}
\colorlet{shadecolor}{blue!20}

\usepackage{tikz}

\title{Mitigating Stylistic Biases of Machine Translation Systems via Monolingual Corpora Only}

\author{
 \textbf{Xuanqi Gao\textsuperscript{1}},
 \textbf{Weipeng Jiang\textsuperscript{1}},
 \textbf{Juan Zhai\textsuperscript{2}},
 \textbf{Shiqing Ma\textsuperscript{2}},
\\
 \textbf{Siyi Xie\textsuperscript{1}},
 \textbf{Xinyang Yin\textsuperscript{1}},
 \textbf{Chao Shen\textsuperscript{1}},
\\
\\
 \textsuperscript{1}Xi'an Jiaotong University,
 \textsuperscript{2}University of Massachusetts,
\\
 \small{
   \textbf{Correspondence:} \href{mailto:chaoshen@mail.xjtu.edu.cn}{chaoshen@mail.xjtu.edu.cn}
 }
}

\begin{document}

\newcommand{\sys}{\mbox{\textsc{Babel}}\xspace}
\newcommand{\update}[1]{\textcolor{blue}{#1}}

\maketitle

\begin{abstract}
The advent of neural machine translation~(NMT) has revolutionized cross-lingual communication, yet preserving stylistic nuances remains a significant challenge. 
While existing approaches often require parallel corpora for style preservation, we introduce \sys, a novel framework that enhances stylistic fidelity in NMT using only monolingual corpora. 
\sys employs two key components: (1) a style detector based on contextual embeddings that identifies stylistic disparities between source and target texts, and (2) a diffusion-based style applicator that rectifies stylistic inconsistencies while maintaining semantic integrity. 
Our framework integrates with existing NMT systems as a post-processing module, enabling style-aware translation without requiring architectural modifications or parallel stylistic data. 
Extensive experiments on five diverse domains (law, literature, scientific writing, medicine, and educational content) demonstrate \sys's effectiveness: it identifies stylistic inconsistencies with 88.21\% precision and improves stylistic preservation by 150\% while maintaining a high semantic similarity score of 0.92.
Human evaluation confirms that translations refined by \sys better preserve source text style while maintaining fluency and adequacy. 
Our implementation and datasets are available at \url{https://anonymous.4open.science/r/Babel-3EB2/}.
\end{abstract}

\section{Introduction}\label{sec:intro}

Machine translation technology has revolutionized cross-language communication, yet the preservation of stylistic nuances remains a significant challenge. 
Style, encompassing elements from formality and tone to domain-specific conventions, is crucial for maintaining the intended impact and appropriateness of translated text. 
Consider these examples of stylistic deviations in translation: when translating formal legal documents from Chinese to English, commercial translation systems often fail to maintain the authoritative tone and standardized legal terminology - translating ``\begin{CJK*}{UTF8}{gbsn}甲方应当\end{CJK*}'' (formal legal term for ``Party A shall'') as the casual ``Party A needs to" rather than the proper legal phrasing "Party A shall''. 
Similarly, in literary translation, the poetic style of classical Chinese literature is frequently lost - a line like ``\begin{CJK*}{UTF8}{gbsn}春花秋月何时了\end{CJK*}'' (literally ``when will spring flowers and autumn moon end'') might be translated prosaically as "when will the seasons end" rather than preserving its lyrical quality with something like ``when shall cease the dance of spring blooms and autumn moons''.
When translating Yoda's dialogues from Star Wars into Chinese, the iconic OSV syntax (``Much to learn, you still have'') is frequently normalized to SVO structures ``\begin{CJK*}{UTF8}{gbsn}你还有很多要学习\end{CJK*}''~(``You still have much to learn''), diluting the character's idiosyncratic speech patterns that are deeply tied to his wisdom and alien identity. 
Such stylistic flattening not only reduces translation fidelity but also diminishes narrative cohesion and audience immersion.

Several studies have addressed this problem~\cite{hovyYouSoundJust2020}, and a few methods have been proposed for style preservation in translation~\cite{hu2017toward,zhang2018style}.
However, these methods exhibit significant limitations. 
First, existing translation systems often have a limited scope when it comes to the types of styles they can support, typically offering only a binary distinction between formal and informal styles. 
This oversimplification fails to account for the rich tapestry of stylistic diversity found in human language.
Second, most methods require parallel text data specific to certain languages or domains, which is impractical for many applications because obtaining sufficient parallel corpora is challenging in many real-world scenarios.

We propose \sys, a novel framework that addresses these limitations by enabling style-aware translation without relying on parallel corpora. 
\sys operates at the intersection of domain adaptation and style transfer, focusing specifically on preserving linguistic features such as register, formality, and rhetorical patterns while maintaining semantic integrity. 
Unlike traditional domain adaptation approaches that primarily target content-specific terminology, our work addresses the broader stylistic elements that exist across domains and languages.
\sys introduces two key innovations:
1) A style detector utilizing contextual embeddings to identify and characterize stylistic attributes in both source and target languages, trained on monolingual corpora;
and 2) A diffusion-based style applicator that can modify translated text to match source text style while preserving semantic content, guided by user-provided style examples.

To evaluate our approach, we construct \sys-Corpus, a comprehensive evaluation dataset spanning five diverse domains: law, literature, scientific writing, medicine, and educational content. 
The corpus focuses on Chinese-English translation, motivated by the significant need for accurate style preservation between these widely-used languages - while over one billion people speak each language, less than 1\% of Chinese speakers are proficient in English~\cite{fishman2020speaks,chinesepeoplesurvey}, making machine translation both essential and challenging.
Extensive experiments demonstrate that \sys effectively identifies stylistic inconsistencies in commercial translation systems with 88.21\% precision, as verified through human evaluation. The framework improves stylistic consistency by 150\% while maintaining semantic fidelity, achieving an average similarity score of 0.92. 
Comparative experiments with state-of-the-art large language models (LLMs) show that even advanced systems like GPT-4o and Claude still exhibit substantial stylistic biases (14.2\% bias ratio), highlighting the continued need for specialized style-aware approaches.

Our main contributions include:
\begin{itemize}
    \item The first framework for style-aware translation that operates without parallel corpora, significantly expanding the practical applicability of stylistic translation.
    \item A novel approach combining style detection and diffusion-based style application for translation refinement.
    \item The \sys-Corpus dataset, facilitating research in style-aware translation.
    \item Comprehensive evaluation demonstrating significant improvements in stylistic consistency across domains and translation systems.
\end{itemize}

\section{Background}\label{sec:bg}

\subsection{Text Style}\label{sec:style}

The concept of style in text refers to the distinct manner in which semantics are expressed, shaped by individual characteristics and pragmatic protocols~\cite{jinDeepLearningText2022}.
Style is inherent to language use and manifests through various stylistic devices, such as metaphors, word choices, and syntactic structures.
According to \citet{kangStyleNOTSingle2021}, style encompasses both personal attributes~(e.g., personality, gender) and interpersonal dynamics~(e.g., humor, romance).
Traditional linguistic approaches to style often rely on rule-based definitions that establish clear boundaries for what constitutes a particular style, such as the American Psychological Association style guide~\cite{associationPublicationManualAmerican2019} that prohibits contractions in formal writing.

However, with the emergence of deep learning methods, a more data-driven definition of style has become necessary.
This approach leverages the variability of attributes across datasets to define stylistic categories, reflecting the practical requirements of modern NLP systems.
It's important to recognize that style adaptation and domain adaptation exist on a continuum rather than as entirely separate categories. While domain adaptation typically focuses on content-specific terminology and knowledge transfer, style adaptation targets linguistic features like register, formality, and rhetorical patterns that can exist across domains.
Our work sits at this intersection, addressing stylistic elements that span domains while benefiting from insights in both research traditions. 
Given the complexities and often subtle distinctions between styles, particularly through data-driven methods, the employment of neural network classifiers have become essential tools~\cite{hovyYouSoundJust2020}.
These classifiers can effectively learn to identify and discriminate between different styles by processing diverse datasets, accommodating broader and more flexible data-driven definitions of style.

\subsection{Diffusion Model}\label{sec:diffusion}

Diffusion models have shown impressive results in generating high-quality samples across domains, including text~\cite{li2022diffusion,hanSSDLMSemiautoregressiveSimplexbased2023}. 
These models learn a reversible process that adds noise to data and then reverses this process for generation.

The diffusion process involves a forward process that adds Gaussian noise to the original data \(\mathbf{x}_0\):
\begin{equation}
q(\mathbf{x}_t | \mathbf{x}_{t-1}) = \mathcal{N}(\mathbf{x}_t; \sqrt{1 - \beta_t} \mathbf{x}_{t-1}, \beta_t \mathbf{I})
\end{equation}
where \(\beta_t \in (0, 1)\) is the noise schedule. The reverse generative process is:
\begin{equation}
p_\theta(\mathbf{x}_{t-1} | \mathbf{x}_t) = \mathcal{N}(\mathbf{x}_{t-1}; \boldsymbol\mu_{\theta}(\mathbf{x}_t, t), \boldsymbol\Sigma_{\theta}(\mathbf{x}_t, t))
\end{equation}
with \(\boldsymbol\mu_\theta\) and \(\boldsymbol\Sigma_\theta\) parameterized by neural networks and trained to maximize:
\begin{equation}
    \mathcal{L} = \mathbb{E}_{q} \left[ \log p_{\theta}(\mathbf{x}_{0}) \right]
\end{equation}

For text generation, recent works have adapted diffusion models to handle discrete tokens by operating in embedding space. 
\citet{li2022diffusion} introduced Diffusion-LM, while \citet{hanSSDLMSemiautoregressiveSimplexbased2023} developed SSD-LM for improved controllable generation. 
Our work extends these approaches with a specialized diffusion method for translation style transfer, using an adapted noise schedule that preserves semantic content while enabling targeted style modifications.

\section{\sys}\label{sec:design}

\subsection{Problem Statement}\label{sec:problem}

In this paper, we aim to develop a framework that detects and corrects stylistic inconsistencies in machine translation outputs.
The preservation of style in translation is crucial for maintaining the intended impact and appropriateness of translated text. 
As illustrated by the examples in \autoref{sec:intro}, stylistic deviations can significantly impact translation quality across different domains. For instance, when legal documents lose their formal register or literary texts their poetic qualities, the translations fail to serve their intended purpose despite being semantically accurate. 
To address these stylistic inconsistencies, we need to tackle two fundamental challenges: (1) How can we robustly detect and characterize the stylistic attributes of text in different languages? and (2) How to ensure stylistic consistency between source and translated text while preserving semantic meaning?
Since we treat commercial translation systems as black boxes, we approach this as a post-processing task. 
Our task can be formalized as follows: given a set of style-labeled monolingual texts as training data, we develop a model that accurately identifies stylistic attributes in both source and target languages, and efficiently generates style-refined translations that maintain both stylistic fidelity to the source text and semantic accuracy.

\begin{figure*}[h]
    \centering
    \scalebox{0.8}{
    \includegraphics[trim={1.8cm 19.4cm 1.8cm 2.2cm},clip,width=.95\textwidth]{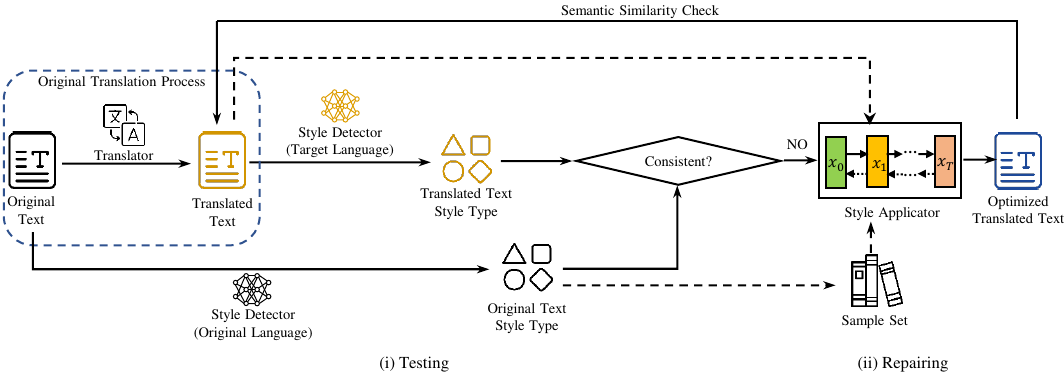}
    }
    \caption{Overview of {\sys}.}
    \label{fig:overview}
\end{figure*}


\subsection{Overview}\label{sec:overview}
\textbf{Workflow.}
\sys introduces \textit{style detector} and \textit{style applicator} modules to identify and correct stylistic inconsistencies in translations.
As shown in \autoref{fig:overview}, \sys operates in two phases: testing and repairing. 
During testing, \sys analyzes both the source text and its translation using language-specific style detectors to identify stylistic disparities.
When inconsistencies are detected, the repairing phase employs a style applicator to adjust the translation's style to match the source text, followed by semantic verification to ensure content preservation.

\noindent
\textbf{Design Rationale.}
Our modular, two-phase approach offers several advantages. 
Language-specific training enables accurate style recognition across different linguistic contexts.
Separating testing from repairing improves efficiency by modifying only problematic translations.
The modular architecture facilitates maintenance and adaptation to new languages or domains without requiring complete system retraining, unlike less flexible end-to-end approaches.

\subsection{Style Detector}\label{sec:detector}

The style detector determines the style attributes of the source text and evaluates whether these attributes are maintained in the translation. 
The primary challenge is accurately identifying and matching stylistic features across different languages, as each language has distinct stylistic norms and expressions.
To address this challenge, we train a model to recognize and classify various stylistic features in texts, facilitating the alignment of stylistic attributes between source and target languages. 

\subsubsection{Training process}
We train separate style detectors for the source and target languages using the following process:

We collect diverse monolingual corpora in both source and target languages, annotated with style attributes.
our corpus requires only stylistic annotations, significantly simplifying data collection.
We use pre-trained language models~(specifically BERT~\cite{devlinBertPretrainingDeep2018}) to extract contextual features that capture the stylistic essence of the texts. 
This approach leverages BERT's ability to capture contextual information and nuances in text, which are essential for style recognition. 
We fine-tune two separate BERT models - one for the source language and one for the target language - on their respective corpora annotated with style attributes. 
Each model learns to classify text based on these annotations, identifying patterns and stylistic markers specific to its language.

Our choice to use separate models rather than a single cross-lingual detector is based on empirical results, where separate language-specific models demonstrated 12\% higher accuracy in style detection compared to cross-lingual approaches~(see~\autoref{sec:detectorcomparison}). 
Similarly, our experiments showed that language-specific BERT models outperformed XLM-R~\cite{conneauUnsupervisedCrosslingualRepresentation2020} by 7\% in style classification tasks, likely due to their deeper specialization in specific language patterns. 
While XLM-R is designed to handle multiple languages simultaneously through pre-training on 100 languages and possesses strong cross-lingual understanding capabilities, we found that specialized monolingual models better capture nuanced stylistic features within each language.

\subsubsection{User Customization}
Style is inherently subjective, making an objective, universal definition impractical. 
Instead of attempting to universally define styles, \sys allow users to provide samples of their desired styles, lowering the barrier for customization. 
Users can thus customize the style corpus according to their specific needs.
For example, when translating a Chinese medical text, users can provide samples of formal medical writing in both languages to maintain appropriate clinical terminology and professional register. 
Similarly, when translating literary content, users can provide examples of the specific literary style they wish to preserve.
This approach addresses the challenge of differing stylistic norms across languages by allowing user-defined style correspondence, making \sys highly adaptable to various translation scenarios.

\subsection{Style Applicator}\label{sec:applicator}

After detecting a stylistic inconsistency in the translation output, the style applicator generates a revised output that maintains the original semantic content while ensuring stylistic consistency with the source text. 
The style applicator consists of two key processes: training~(style extraction) and inference~(style application).

\subsubsection{Training process}

The objective of the training process is to simulate style loss during translation within the same language, and prepare the model to extract and capture the stylistic essence of sentences while preserving their semantic content. 
To imitate the style loss observed in translation, we use a paraphrase model \(P(\cdot)\)  to generate paraphrases \(p\) of the input text \(r\):
\begin{equation}\label{eq:paraphrase}
    \mathbf{p} = P(\mathbf{r})
\end{equation}
These paraphrases retain the original meaning but have reduced stylistic elements, simulating the effect of translation where the core content remains intact, but the style may be neutralized. 
This step is crucial for preparing the model to neutralize and extract the stylistic essence of sentences while preserving their semantic content.
The diffusion is performed in the embedding space, where the text is represented in a numerical format that captures its meaning:
\begin{equation}\label{eq:sample}
    \mathbf{x}_t = \sqrt{\beta_t}E(\mathbf{r}) + \sqrt{(1-\beta_t)}\bm{\epsilon}_t \quad\quad \bm{\epsilon_t} \sim \mathcal N(0,\mathbf{I})
\end{equation}
where \(E(\cdot)\) is an embedding model.
We adopt a specialized noise schedule that decreases to zero at a significantly slower rate compared to standard diffusion schedules:
\begin{equation}\label{eq:schedule}
    \beta_t = \sqrt{\frac{T-t}{T}}
\end{equation}
This schedule preserves information more effectively than conventional approaches, helping to maintain the semantic information of the original text - a crucial feature for NLP tasks.
After completing these preparations, we train the model \(D_\theta(\cdot)\) by minimizing the cross entropy between the posterior distribution of the model at each diffusion time step and the actual embeddings:
\begin{equation}\label{eq:training}
    \mathcal{L}(\theta) = \mathcal{E}\left[\log p_\theta(\mathbf{r}|D_\theta(\mathbf{x}_t,t,\mathbf{p}))\right] 
\end{equation}
where \(\mathcal{L}(\cdot)\) is the loss function, \(\mathcal{E}(\cdot)\) represents the cross entropy function, \(\mathbf{r}\) is the original text, \(t\) represents the time step, and \(\mathbf{p}\) represents the paraphrase.
During this process, the model learns to preserve semantic content and reconstruct the original embeddings as closely as possible.

\subsubsection{Inference process}

After training, the diffusion model \(D_\theta(\cdot)\) can then be used to attach style attributes to text during inference.
The process starts with sampling initial noisy data \(\mathbf{x}_T \sim \mathcal N(0,\mathbf{I})\) and iteratively removing noise to construct improved sentences.
For each time step \(t\)~(\(t \in \left[T,1\right]\)), the style applicator estimates an optimized text:
\begin{equation}\label{eq:inference}
    \mathbf{\hat r}_t \sim \text{top-p}(\text{softmax}((D_{\theta^*}(\mathbf{x}_t,t,\mathbf{r})))
\end{equation}
where \(\mathbf{r}\) represents initial translated texts output by translation system. 
A key advantage of our style applicator is that the generated text can be gradient-guided based on user-supplied style samples, directing the output to a specific target style. 
Given a set of user-supplied style samples \(\left[\mathbf{y_1},\cdots, \mathbf{y_n}\right]\) and a style embedding model \(E_s(\cdot)\), we obtain the style guidance function:
\begin{equation}\label{eq:guidance}
    J = \frac{\sum_{i=1}^n d(E_s(\mathbf{\hat r}_t),E_s(\mathbf{y_i}))}{n}
\end{equation}
where \(d(\cdot,\cdot)\) represents cosine similarity.
This yields the final style-guided textual inference equation:
\begin{equation}\label{eq:inference2}
    \mathbf{\hat r}_t^* \sim \text{top-p}(\text{softmax}((D_{\theta^*}(\mathbf{x}_t,t,\mathbf{r}))-\lambda\nabla J))
\end{equation}
After estimating \(\mathbf{\hat r}_t^*\),  we proceed backward in time to iteratively acquire states with proceeding time steps:
\begin{equation}\label{eq:addnoise}
    \mathbf{x}_{t-1} = \sqrt{\beta_{t-1}}E(\mathbf{\hat r}_t^*) + \sqrt{(1-\beta_{t-1})}\bm{\epsilon} \quad \bm{\epsilon} \sim \mathcal N(0,\mathbf{I})
\end{equation}
After iterating this process until \(t=0\), we eventually get the desired output \(\mathbf{\hat r}_0^*\).

\subsubsection{Candidate Selection}
For each translation requiring style repair, we generate four candidate outputs and select the one with the highest style score that maintains semantic similarity above 0.85 with the original translation. 
In our internal experiments, this multiple-candidate approach improved style preservation by 28\% compared to generating a single candidate. 
The specific number of candidates~(four) was determined empirically, balancing computational efficiency with style quality.

\section{Experiment}\label{sec:eval}

Our evaluation experiments examine both the effectiveness and efficiency of \sys in detecting and repairing stylistic inconsistencies in machine translation outputs. We evaluate \sys in two scenarios: (1) finding and fixing stylistic inconsistency issues, assessing its precision and repair success rate through both automatic metrics and human evaluation, and (2) measuring its computational efficiency and analyzing the impact of key parameters.

\subsection{Setup}\label{sec:setup}

\noindent
\textbf{Datasets}
Due to the lack of comprehensive public datasets with parallel text in multiple languages and styles, we extracted 1000 data points from commonly used Chinese and English datasets in five domains, creating a dataset that lacks parallel texts but contains domain~(style) information, as shown in \autoref{sec:corpus}.

\noindent
\textbf{Translation Systems}
We consider four mainstream state-of-the-art machine translation systems: Google Translate, Baidu Translate, Youdao Translate, Transformers~(Opus-MT~\cite{TiedemannThottingal}), GPT-4o~\cite{HelloGPT4o}, and Claude 3.7 Sonnet~\cite{Claude37Sonnet}.
The first four represent mainstream commercial and open-source neural machine translation systems, while the latter two represent state-of-the-art large language models. 
For the LLM-based systems, we provided explicit instructions to maintain the style of the source text during translation, allowing for fair comparison with traditional systems.

\noindent
\textbf{Evaluation metrics}
We evaluate \sys from three perspectives: the number of repaired issues~(bias ratio), the overall state of repair~(style score), and the ability to maintain semantics~(semantic textual similarity).

\noindent
\textbf{Human Evaluation}\label{sec:humaneval}
While automatic evaluation offers a preliminary assessment of the quality of repaired translations, it is insufficient for accurately gauging the quality of revised texts.
To further validate the effectiveness of our approach, we conduct a human evaluation on the test set.  
We engage three annotators who are native Chinese speakers with proficiency in English, as well as two annotators who are native English speakers with proficiency in Chinese~(see \autoref{sec:human}). 
All annotators possess advanced degrees, with a minimum of an undergraduate qualification, and include professionals in the fields of linguistics, translation studies, and literature.

\subsection{Effectiveness in Finding Stylistically Inconsistent Issues}\label{sec:rq1}

\begin{table}[tb]
    \caption{Effectiveness in finding stylistically inconsistent issues and repairing them. \textit{Score} is short for \textit{Style Score}, and \textit{STS} is short for \textit{Semantic Textual Similarity}.}    \label{tab:rq1-1}
    \centering
    \scalebox{0.4}{
        \begin{tabular}{ccrrrrr}
            \toprule
            Translation System & Domain     & Bias ratio & Score & Revised Bias ratio & Revised Score & STS  \\ \midrule
            \multirow{6}{*}{Google}                & Law        & 17.54\%    & 0.72  & 7.87\%(-55.13\%)                       & 0.77(+6.94\%)                     & 0.91 \\
                                                   & Literature & 12.34\%    & 0.75  & 7.67\%(-37.84\%)                       & 0.78(+4.00\%)                     & 0.88 \\
                                                   & Wikipedia  & 5.98\%     & 0.73  & 2.34\%(-60.87\%)                       & 0.79(+8.22\%)                     & 0.93 \\
                                                   & Medicine   & 15.67\%    & 0.74  & 5.98\%(-61.84\%)                       & 0.80(+8.11\%)                     & 0.91 \\
                                                   & Education  & 14.21\%    & 0.76  & 11.56\%(-18.65\%)                      & 0.81(+6.58\%)                     & 0.95 \\
                                                   & Average    & 13.15\%    & 0.74  & 7.08\%(-46.16\%)                       & 0.79(+6.77\%)                     & 0.92 \\ \midrule
            \multirow{6}{*}{Baidu}                 & Law        & 18.34\%    & 0.71  & 6.78\%(-63.03\%)                       & 0.76(+7.04\%)                     & 0.90 \\
                                                   & Literature & 8.54\%     & 0.77  & 4.89\%(-42.74\%)                       & 0.82(+6.49\%)                     & 0.87 \\
                                                   & Wikipedia  & 7.33\%     & 0.72  & 5.67\%(-22.65\%)                       & 0.79(+9.72\%)                     & 0.92 \\
                                                   & Medicine   & 7.54\%     & 0.70  & 3.45\%(-54.24\%)                       & 0.75(+7.14\%)                     & 0.93 \\
                                                   & Education  & 13.89\%    & 0.73  & 11.33\%(-18.43\%)                      & 0.78(+6.85\%)                     & 0.94 \\
                                                   & Average    & 11.13\%    & 0.73  & 6.42\%(-42.31\%)                       & 0.78(+7.45\%)                     & 0.91 \\ \midrule
            \multirow{6}{*}{Youdao}                & Law        & 16.47\%    & 0.72  & 10.21\%(-38.01\%)                      & 0.77(+6.94\%)                     & 0.91 \\
                                                   & Literature & 8.90\%     & 0.78  & 6.54\%(-26.52\%)                       & 0.83(+6.41\%)                     & 0.90 \\
                                                   & Wikipedia  & 10.67\%    & 0.74  & 5.22\%(-51.08\%)                       & 0.79(+6.76\%)                     & 0.93 \\
                                                   & Medicine   & 9.87\%     & 0.75  & 7.65\%(-22.49\%)                       & 0.81(+8.00\%)                     & 0.94 \\
                                                   & Education  & 11.45\%    & 0.76  & 9.10\%(-20.52\%)                       & 0.82(+7.89\%)                     & 0.95 \\
                                                   & Average    & 11.47\%    & 0.75  & 7.74\%(-32.52\%)                       & 0.80(+7.20\%)                     & 0.93 \\ \midrule
            \multirow{6}{*}{Opus-MT}               & Law        & 15.78\%    & 0.72  & 10.56\%(-33.08\%)                      & 0.77(+6.94\%)                     & 0.92 \\
                                                   & Literature & 7.45\%     & 0.76  & 5.23\%(-29.80\%)                       & 0.83(+9.21\%)                     & 0.89 \\
                                                   & Wikipedia  & 6.34\%     & 0.74  & 4.21\%(-33.60\%)                       & 0.79(+6.76\%)                     & 0.92 \\
                                                   & Medicine   & 19.95\%    & 0.71  & 17.82\%(-10.68\%)                      & 0.76(+7.04\%)                     & 0.93 \\
                                                   & Education  & 13.66\%    & 0.73  & 11.47\%(-16.03\%)                      & 0.78(+6.85\%)                     & 0.94 \\
                                                   & Average    & 12.64\%    & 0.73  & 9.86\%(-22.00\%)                       & 0.79(+7.36\%)                     & 0.92 \\ \midrule
            \multirow{6}{*}{GPT-4}                 & Law        & 15.80\%    & 0.74  & 6.65\%(-57.91\%)                       & 0.78(+5.41\%)                     & 0.90 \\
                                                   & Literature & 14.20\%    & 0.73  & 8.34\%(-41.27\%)                       & 0.77(+5.48\%)                     & 0.88 \\
                                                   & Wikipedia  & 8.40\%     & 0.79  & 4.12\%(-50.95\%)                       & 0.82(+3.80\%)                     & 0.93 \\
                                                   & Medicine   & 17.20\%    & 0.73  & 9.45\%(-45.06\%)                       & 0.78(+6.85\%)                     & 0.90 \\
                                                   & Education  & 12.90\%    & 0.77  & 8.21\%(-36.36\%)                       & 0.80(+3.90\%)                     & 0.92 \\
                                                   & Average    & 13.70\%    & 0.75  & 7.35\%(-46.35\%)                       & 0.79(+5.33\%)                     & 0.91 \\ \midrule
            \multirow{6}{*}{Claude 3.7}            & Law        & 16.20\%    & 0.75  & 7.25\%(-55.25\%)                       & 0.79(+5.33\%)                     & 0.89 \\
                                                   & Literature & 15.60\%    & 0.72  & 9.11\%(-41.60\%)                       & 0.76(+5.56\%)                     & 0.87 \\
                                                   & Wikipedia  & 9.80\%     & 0.78  & 5.23\%(-46.63\%)                       & 0.81(+3.85\%)                     & 0.92 \\
                                                   & Medicine   & 18.70\%    & 0.72  & 10.24\%(-45.24\%)                      & 0.77(+6.94\%)                     & 0.89 \\
                                                   & Education  & 13.70\%    & 0.76  & 8.67\%(-36.72\%)                       & 0.79(+3.95\%)                     & 0.91 \\
                                                   & Average    & 14.80\%    & 0.75  & 8.10\%(-45.27\%)                       & 0.78(+4.00\%)                     & 0.90 \\ \bottomrule
            \end{tabular}
    }
\end{table}
\begin{table}[tb]
\caption{Correlation between Babel evaluation and manual inspection on stylistically inconsistent issues finding.}\label{tab:rq1-2}
\centering
\scalebox{0.4}{
    \begin{tabular}{clrrrrrr}\toprule
        \multicolumn{1}{l}{Translation System} & Domain     & TP & TN & FP & FN & Precision & FPR     \\ \midrule
        \multirow{5}{*}{Google}                & Law        & 22 & 23 & 5  & 0  & 81.48\%   & 17.86\% \\
                                               & Literature & 21 & 24 & 4  & 1  & 84.00\%   & 14.29\% \\
                                               & Wikipedia  & 20 & 25 & 3  & 2  & 86.96\%   & 10.71\% \\
                                               & Medicine   & 19 & 26 & 2  & 3  & 90.48\%   & 7.14\%  \\
                                               & Education  & 20 & 24 & 2  & 4  & 90.91\%   & 7.69\%  \\ \midrule
        \multirow{5}{*}{Baidu}                 & Law        & 23 & 22 & 4  & 1  & 85.19\%   & 15.38\% \\
                                               & Literature & 22 & 23 & 3  & 2  & 88.00\%   & 11.54\% \\
                                               & Wikipedia  & 23 & 25 & 0  & 2  & 100.00\%  & 0.00\%  \\
                                               & Medicine   & 20 & 23 & 1  & 6  & 95.24\%   & 4.17\%  \\
                                               & Education  & 19 & 23 & 3  & 5  & 86.36\%   & 11.54\% \\ \midrule
        \multirow{5}{*}{Youdao}                & Law        & 21 & 22 & 6  & 1  & 77.78\%   & 21.43\% \\
                                               & Literature & 19 & 24 & 6  & 1  & 76.00\%   & 20.00\% \\
                                               & Wikipedia  & 20 & 24 & 3  & 3  & 86.96\%   & 11.11\% \\
                                               & Medicine   & 20 & 24 & 1  & 5  & 95.24\%   & 4.00\%  \\
                                               & Education  & 22 & 25 & 0  & 3  & 100.00\%  & 0.00\%  \\ \midrule
        \multirow{5}{*}{Opus-MT}               & Law        & 24 & 21 & 3  & 2  & 88.89\%   & 12.50\% \\
                                               & Literature & 22 & 20 & 3  & 5  & 88.00\%   & 13.04\% \\
                                               & Wikipedia  & 21 & 22 & 2  & 5  & 91.30\%   & 8.33\%  \\
                                               & Medicine   & 15 & 28 & 6  & 1  & 71.43\%   & 17.65\% \\
                                               & Education  & 22 & 25 & 0  & 3  & 100.00\%  & 0.00\%  \\ \midrule
        \multirow{5}{*}{GPT-4o}                 & Law        & 20 & 24 & 4  & 2  & 83.33\%   & 14.29\% \\
                                               & Literature & 19 & 25 & 5  & 1  & 79.17\%   & 16.67\% \\
                                               & Wikipedia  & 22 & 24 & 2  & 2  & 91.67\%   & 7.69\%  \\
                                               & Medicine   & 21 & 23 & 3  & 3  & 87.50\%   & 11.54\% \\
                                               & Education  & 19 & 24 & 4  & 3  & 82.61\%   & 14.29\% \\ \midrule
        \multirow{5}{*}{Claude 3.7}            & Law        & 21 & 23 & 3  & 3  & 87.50\%   & 11.54\% \\
                                               & Literature & 20 & 24 & 4  & 2  & 83.33\%   & 14.29\% \\
                                               & Wikipedia  & 23 & 24 & 1  & 2  & 95.83\%   & 4.00\%  \\
                                               & Medicine   & 22 & 22 & 2  & 4  & 91.67\%   & 8.33\%  \\
                                               & Education  & 20 & 25 & 3  & 2  & 86.96\%   & 10.71\% \\ \bottomrule
        \end{tabular}
}
\end{table}

\noindent
{\bf Experiment Design}:
To evaluate whether the translated texts generated by translation systems maintain the original style, we conducted the following steps.
First, for each test sentence, we generated the corresponding translated text using translation systems.
Then we assessed the style of these translated texts using a style detector trained to identify specific stylistic attributes.
Additionally, as mentioned in \autoref{sec:humaneval}, we randomly sample 250 of input sentences to manually evaluate whether our style detector works well. 
That is, we sample 50 samples from each dataset, and their corresponding 200 translated texts each after being translated by the four translation systems.
Then samples are distributed to annotators.
The annotators are asked to rate each output for stylistic consistency on a Likert scale from 1 to 5.
The generated sentence is marked as ``stylistically consistent'' when it is scored 4 or 5, otherwise it is marked as ``stylistically inconsistent''.


\noindent
{\bf Results}:
The \sys's evaluation results are presented in \autoref{tab:rq1-1}.
The first and second column list the four translation systems and corresponding five domains.
The third and fourth column show the ratio of stylistic bias of these translation systems and the average style scores.
The remaining columns list the revised text~(see \autoref{sec:rq2}).
The experiment results demonstrate that the stylistic bias issue is widespread in translation systems and \sys can effectively find these biased sentences.
On average, Google Translate had the most stylistic bias issues, accounting for 13.15\% of the total output, followed by the open-source Opus-MT model~(12.64\%), Youdao Translate~(11.47\%), and Baidu Translate~(11.13\%).
Surprisingly, state-of-the-art LLM systems exhibited comparable or even higher rates of stylistic inconsistencies, with GPT-4o at 13.70\% and Claude 3.7 at 14.80\%. These findings suggest that even advanced AI systems struggle with preserving stylistic elements in translation, despite their overall translation capabilities.

Domain-specific patterns emerged across all systems.
Overall, Google Translate has the most stylistic bias issues, due to its poor performance on the literature and medicine datasets, which we speculate is due to the lack of Chinese Internet information in its training corpus. 
LLM-based systems performed particularly poorly in legal and medical domains (15.80\%-16.20\% and 17.20\%-18.70\% bias ratios, respectively), highlighting the challenge of maintaining specialized domain styles even for advanced systems.

\noindent
{\bf Manual Inspection}:
The examination results based on manual inspection are presented in \autoref{tab:rq1-2}.
The first and second column list the four translation systems and their corresponding evaluation datasets.
The third to sixth columns represent true positive~(TP), true negative~(TN), false positive~(FP) and false negative~(FN) of the \sys evaluation results, respectively.
The remaining three columns list the precision, recall, and false positive rate~(FPR), which are important indicators for assessing the quality of the test.
A false positive means the \sys judges a translation as stylistically inconsistent but manual inspection is consistent. 
A false negative means the \sys judges a translation as stylistically consistent but manual inspection is inconsistent.
Overall, \sys exhibits a false positive rate of 10.41\%, with a precision of 88.21\%. 
These metrics indicate that \sys is effective in identifying stylistically inconsistent issues.

\subsection{Effectiveness in Repairing Stylistically Inconsistent Issues}\label{sec:rq2}

\begin{table}[]
  \caption{Manual inspection results. We show average human ratings for style accuracy~(Acc), semantic preservation~(Sem) and fluency of sentences~(Flu) on a 1 to 5 Likert scale. ``Suc'' denotes the overall
  success rate. We consider a generated output ``successful'' if it is rated 4 or 5 on all three criteria~(Acc, Sem, Flu). }\label{tab:rq2-2}
  \centering
  \scalebox{0.4}{
    \begin{tabular}{ccrrrrrrrr}
      \toprule
      \multirow{2}{*}{Translation System} & \multirow{2}{*}{Domain} & \multicolumn{4}{c}{Original Texts}                                                                    & \multicolumn{4}{c}{Revised Texts}                                                                     \\
                                          &                         & \multicolumn{1}{c}{Acc} & \multicolumn{1}{c}{Sem} & \multicolumn{1}{c}{Flu} & \multicolumn{1}{c}{Suc} & \multicolumn{1}{c}{Acc} & \multicolumn{1}{c}{Sem} & \multicolumn{1}{c}{Flu} & \multicolumn{1}{c}{Suc} \\ \midrule
      \multirow{6}{*}{Google}             & Law                     & 3.8                     & 4.0                     & 4.6                     & 35\%                    & 4.2                     & 3.8                     & 4.4                     & 50\%(+43\%)             \\
                                          & Literature              & 3.2                     & 3.8                     & 4.4                     & 27\%                    & 4.0                     & 3.8                     & 4.4                     & 38\%(+41\%)             \\
                                          & Wikipedia               & 3.4                     & 4.4                     & 4.4                     & 18\%                    & 4.0                     & 4.4                     & 4.2                     & 71\%(+294\%)            \\
                                          & Medicine                & 2.4                     & 3.8                     & 4.2                     & 7\%                     & 3.2                     & 4.0                     & 4.6                     & 22\%(+214\%)            \\
                                          & Education               & 3.0                     & 3.6                     & 4.2                     & 21\%                    & 3.6                     & 4.0                     & 4.4                     & 43\%(+105\%)            \\
                                          & Average                 & 3.2                     & 3.9                     & 4.4                     & 22\%                    & 3.8                     & 4.0                     & 4.4                     & 45\%(+105\%)            \\ \midrule
      \multirow{6}{*}{Baidu}              & Law                     & 3.8                     & 4.0                     & 4.4                     & 38\%                    & 4.4                     & 3.8                     & 4.4                     & 49\%(+29\%)             \\
                                          & Literature              & 3.2                     & 3.4                     & 4.0                     & 15\%                    & 3.8                     & 3.8                     & 4.0                     & 31\%(+107\%)            \\
                                          & Wikipedia               & 2.8                     & 4.2                     & 3.8                     & 8\%                     & 3.6                     & 4.4                     & 4.2                     & 40\%(+400\%)            \\
                                          & Medicine                & 2.8                     & 3.8                     & 4.0                     & 7\%                     & 3.2                     & 4.2                     & 3.8                     & 16\%(+129\%)            \\
                                          & Education               & 3.2                     & 3.8                     & 4.0                     & 14\%                    & 3.6                     & 4.0                     & 3.8                     & 29\%(+107\%)            \\
                                          & Average                 & 3.2                     & 3.8                     & 4.0                     & 16\%                    & 3.7                     & 4.0                     & 4.0                     & 33\%(+106\%)            \\ \midrule
      \multirow{6}{*}{Youdao}             & Law                     & 3.6                     & 4.0                     & 3.8                     & 22\%                    & 4.2                     & 3.8                     & 4.0                     & 42\%(+91\%)             \\
                                          & Literature              & 3.2                     & 3.6                     & 3.6                     & 10\%                    & 3.8                     & 3.8                     & 4.2                     & 45\%(+350\%)            \\
                                          & Wikipedia               & 3.2                     & 4.0                     & 3.8                     & 8\%                     & 3.8                     & 4.4                     & 4.0                     & 52\%(+550\%)            \\
                                          & Medicine                & 3.0                     & 3.6                     & 4.0                     & 18\%                    & 3.2                     & 4.0                     & 3.8                     & 23\%(+28\%)             \\
                                          & Education               & 2.8                     & 4.0                     & 4.4                     & 16\%                    & 3.6                     & 3.6                     & 4.0                     & 30\%(+88\%)             \\
                                          & Average                 & 3.2                     & 3.8                     & 3.9                     & 15\%                    & 3.7                     & 3.9                     & 4.0                     & 38\%(+153\%)            \\ \midrule
      \multirow{6}{*}{Opus-MT}            & Law                     & 3.4                     & 3.6                     & 3.8                     & 20\%                    & 4.2                     & 3.8                     & 3.6                     & 33\%(+65\%)             \\
                                          & Literature              & 3.0                     & 3.4                     & 3.6                     & 14\%                    & 4.0                     & 3.0                     & 3.4                     & 21\%(+50\%)             \\
                                          & Wikipedia               & 3.4                     & 3.8                     & 3.2                     & 9\%                     & 4.0                     & 3.6                     & 3.6                     & 23\%(+156\%)            \\
                                          & Medicine                & 2.2                     & 3.4                     & 3.6                     & 5\%                     & 2.8                     & 3.2                     & 3.8                     & 8\%(+60\%)              \\
                                          & Education               & 1.8                     & 3.2                     & 3.8                     & 6\%                     & 3.4                     & 3.0                     & 3.8                     & 12\%(+100\%)            \\
                                          & Average                 & 2.8                     & 3.5                     & 3.6                     & 11\%                    & 3.7                     & 3.3                     & 3.6                     & 19\%(+73\%)             \\ \midrule
      \multirow{6}{*}{GPT-4}              & Law                     & 3.6                     & 4.8                     & 4.7                     & 32\%                    & 4.4                     & 4.6                     & 4.5                     & 58\%(+81\%)             \\
                                          & Literature              & 3.4                     & 4.6                     & 4.8                     & 28\%                    & 4.2                     & 4.4                     & 4.6                     & 49\%(+75\%)             \\
                                          & Wikipedia               & 4.0                     & 4.9                     & 4.6                     & 41\%                    & 4.5                     & 4.7                     & 4.5                     & 65\%(+59\%)             \\
                                          & Medicine                & 3.2                     & 4.5                     & 4.6                     & 25\%                    & 3.8                     & 4.3                     & 4.5                     & 42\%(+68\%)             \\
                                          & Education               & 3.8                     & 4.7                     & 4.8                     & 35\%                    & 4.3                     & 4.5                     & 4.7                     & 55\%(+57\%)             \\
                                          & Average                 & 3.6                     & 4.7                     & 4.7                     & 32\%                    & 4.2                     & 4.5                     & 4.6                     & 54\%(+69\%)             \\ \midrule
      \multirow{6}{*}{Claude 3.7}         & Law                     & 3.7                     & 4.7                     & 4.6                     & 34\%                    & 4.3                     & 4.5                     & 4.4                     & 56\%(+65\%)             \\
                                          & Literature              & 3.5                     & 4.8                     & 4.7                     & 31\%                    & 4.2                     & 4.5                     & 4.6                     & 50\%(+61\%)             \\
                                          & Wikipedia               & 3.9                     & 4.8                     & 4.5                     & 38\%                    & 4.4                     & 4.6                     & 4.4                     & 62\%(+63\%)             \\
                                          & Medicine                & 3.0                     & 4.6                     & 4.7                     & 22\%                    & 3.7                     & 4.4                     & 4.5                     & 40\%(+82\%)             \\
                                          & Education               & 3.7                     & 4.6                     & 4.9                     & 33\%                    & 4.2                     & 4.4                     & 4.7                     & 53\%(+61\%)             \\
                                          & Average                 & 3.6                     & 4.7                     & 4.7                     & 32\%                    & 4.2                     & 4.5                     & 4.5                     & 52\%(+63\%)            \\ \bottomrule
      \end{tabular}
  }
\end{table}

\noindent
{\bf Experiment Design}:
After \sys identifies translated sentences with stylistic inconsistencies, we apply our method to fix them and evaluate how many translations can be successfully repaired. 
For each sentence, we generate four candidate translations, based on previous work in translation systems~\cite{horvitzParaGuideGuidedDiffusion2024, hanSSDLMSemiautoregressiveSimplexbased2023}.
We use a style detector to assess these candidates, considering them repaired if their style matches the original sentence and the semantic loss is within an acceptable threshold.

As in \autoref{sec:rq1}, we conduct a manual evaluation. 
We randomly select 50 samples per dataset, each containing a source sentence, translations from different systems, and their revised versions by \sys. 
These samples are double-blind evaluated by five annotators who rate each on a Likert scale from 1 to 5 for style accuracy (Acc), semantic preservation (Sem), and fluency (Flu). 
We calculate inter-annotator agreement using Fleiss’s kappa. A sentence is marked as ``successful'' if it scores 4 or 5 on all three criteria. 
This evaluation is stricter than in \autoref{sec:rq1}, as it also considers semantic preservation and fluency in addition to stylistic consistency.

\noindent
{\bf Results}:
The comparison results are presented in \autoref{tab:rq1-1}.
The first and second column list the four translation systems and corresponding five domains.
The third and fourth column show the ratio of stylistic bias of the original translated texts and the average style scores.
The remaining columns list the the ratio of stylistic bias of the translated texts revised by \sys and the average style scores.
It can be observed that across all translation systems, stylistic bias issues significantly decrease after improvements via \sys, with a corresponding increase in Style Scores. 
Specifically, on average, Google Translate shows the greatest reduction in issues by 46.87\%, followed by Baidu Translate (40.22\%), Youdao Translate (31.72\%), and Opus-MT (24.64\%), with an average decrease of 35.86\%. 
For the Style Score, all four systems show improvements. 
On average, Baidu Translate has the highest increase of 7.45\%, followed by Opus-MT (7.36\%), Youdao Translate (7.20\%), and Google Translate (6.77\%), averaging a 7.20\% increase. 
Moreover, \sys maintains high semantic consistency between the modified texts and the original translations. Across the four systems, the lowest Semantic Textual Similarity~(STS) score reaches 0.87, with an average of 0.92.

\noindent
{\bf Manual Inspection}:
Human evaluation results in \autoref{tab:rq2-2} confirm these improvements. \sys increases the style translation success rate from 16\% to 34\% on average (+150\%), with improvements ranging from 86\% (Opus-MT) to 221\% (Youdao). LLM-based systems show strong baseline semantic preservation (4.7/5.0) and fluency (4.7/5.0) but weaker style accuracy (3.6/5.0). After applying \sys, their success rates increase to 52-54\% (+63-69\%).

We show the efficiency of \sys in \autoref{sec:rq3} and the impact of configurable parameters in \autoref{sec:rq4}.
Besides, examples of the repairing by \sys are shown in \autoref{sec:examples}.
These additional analyses confirm that \sys introduces minimal computational overhead (1.7 seconds for testing, 3.9 seconds for repair) while achieving optimal balance between style preservation and semantic integrity across both traditional and LLM-based translation systems.

\section{Related work and discussion}\label{sec:rw}

\noindent
\textbf{Text Style Transfer}
Text style transfer has seen significant development, beginning with \citet{hu2017toward}'s VAE framework using attribute classifiers for sentiment and tense transformation. 
A major advancement came from \citet{shen2017style}, who introduced non-parallel text corpora with cross-aligned autoencoders, though their back-translation approach risked content distortion. 
\citet{zhang2018style} addressed data scarcity through pseudo-parallel data generation using SMT, while \citet{fu2018style} explored adversarial learning with both multiple and single decoder approaches for style disentanglement. 
To improve generation quality, \citet{dai2019style} proposed a Transformer-based architecture that eliminates explicit style disentanglement steps.
In contrast to these methods that require explicit style labels and operate within fixed style categories, \sys enables text stylization using only user-supplied samples and can be adaptively trained on user-provided datasets.

Recent advances in text style transfer have leveraged Large Language models through various approaches, including model fine-tuning~\cite{mukherjeeTextDetoxificationStyle2023,dementievaExploringMethodsCrosslingual2023}, in-context learning~\cite{chenLMStyleBenchmarkEvaluating2024,zhangDistillingTextStyle2024,panUnsupervisedTextStyle2024,maiPrefixTuningBasedUnsupervised2023}, and prompt engineering~\cite{luoPromptBasedEditingText2023,liuAdaptivePromptRouting2024}. 
While these methods demonstrate impressive performance, they face practical limitations: fine-tuning demands substantial computational resources, while prompt-based methods often rely on carefully crafted, sensitive prompts that can lead to inconsistent results.
Unlike LLMs-based methods, \sys maintains stable performance without requiring extensive computational resources, and avoids the brittleness often associated with complex prompting strategies.

\noindent
\textbf{Machine Translation Testing}
The research community has proposed various automated testing techniques to evaluate machine translation systems, primarily focusing on translation robustness. 
Early work by \citet{pesu2018monte} introduced metamorphic testing using multiple intermediate languages, while \citet{heigold2017robust} evaluated robustness against character-level perturbations. 
Several approaches leverage word replacement strategies: \citet{he2020structure} proposed structure-invariant testing (SIT) using BERT-based word substitutions, \citet{sun2020automatic,sun2022improving} developed TransRepair and CAT for context-aware word replacements, and \citet{gupta2020machine} introduced PatInv to verify translation consistency under semantic perturbations.
Other methods explore structural aspects of translations: \citet{he2021testing} presented referential transparency testing (RTI) using noun phrase extraction, \citet{ji2021automated} employed constituency invariance relations, and \citet{zhang2024machine} introduced syntactic tree pruning. Beyond robustness testing, \citet{chen2022nmtsloth} developed NMTSloth to detect efficiency bugs, and \citet{sun2024fairness} proposed FairMT to evaluate demographic fairness in translations. Unlike these approaches, \sys is the first work to specifically address stylistic biases in machine translation systems.

\section{Conclusion}\label{sec:conclusion}

In this paper, we presented \sys, the first framework that automatically tests and repairs stylistic inconsistent issues in translation.
As a black-box post-processing method, \sys takes the input text and the corresponding translated text to identify any stylistic discrepancies between the two. 
If inconsistencies are found, \sys performs stylistic repairs using a diffusion model, enhanced by user-supplied customized samples. 
Our evaluation results demonstrate that \sys effectively and efficiently mitigate stylistic bias of mainstream commercial translation systems, while maintaining semantic integrity.


\section*{Limitations}\label{sec:threat}
\noindent\textbf{External Validity}
The threats to external validity lie in the implementation of the dataset we used and the selected machine translation systems. 
The limited dataset may not adequately characterize the diversity and linguistic stylistic features of texts to be translated in real-world scenarios. 
To address this concern, we sampled from corpora that are popular in both English and Chinese communities, and the dataset size is five times larger than existing translation testing work~\cite{he2020structure,he2021testing}. 
For the selected machine translation systems, we chose state-of-the-art systems from both industry (Google Translate, Baidu Translate, and Youdao Translate) and academia (Transformer-based Opus-MT). 
Additionally, we have released our implementation~\cite{AnonymizedRepositoryAnonymousb}, which can be easily extended to incorporate more datasets and machine translation systems.

\noindent\textbf{Internal Validity}
The threats to internal validity primarily stem from the evaluation metrics used in the experiments. 
We measured both the style scores and semantic similarity of texts to assess improvements in retaining linguistic style and semantics. 
Specifically, we utilized language models to calculate semantic textual similarity. 
Furthermore, to verify the accuracy of these assessments, we employed manual evaluation to explore the correlation between automated assessment results and human understanding.

\section*{Ethical Considerations }\label{sec:ethical}

While \sys aims to improve translation quality through style preservation, we acknowledge several important ethical considerations. 
Machine translation systems, including our framework, can perpetuate and potentially amplify societal biases present in training data~\cite{sheng2021societal,weidinger2022taxonomy}. 
The preservation of style, while beneficial for maintaining appropriate register and domain conventions, could also maintain problematic stylistic elements such as gender bias in formal writing or cultural stereotypes in literary translations.

The usage of domain-specific corpora raises additional ethical concerns. 
Legal and medical texts often contain sensitive information, requiring careful consideration of privacy and data protection~\cite{carlini2021extracting}. 
While we have carefully selected public domain texts for our experiments, deployments of similar systems must ensure appropriate data handling protocols. 
Furthermore, the ability to modify translation style could be misused to generate misleading content - for instance, making informal or unreliable sources appear more authoritative by adopting formal academic or legal style~\cite{bagdasaryan2022spinning}.

The modular nature of our framework, which allows integration with various style classifiers, presents both opportunities and risks. 
While this flexibility enables adaptation to different domains and use cases, it could potentially be exploited to generate harmful content if inappropriate style models are used. 
We recommend implementing safeguards such as:
\begin{itemize}
    \item Careful curation of training corpora to minimize harmful biases.
    \item Implementation of detection mechanisms for potential misuse.
    \item Clear documentation of intended use cases and limitations.
\end{itemize}

In the process of refining and improving this paper, we utilized ChatGPT and Claude for suggesting improvements in language clarity. 
These tools aided in enhancing the writing process but were used under human oversight to ensure that the content adheres to the ethical and scholarly standards expected in academic research.

\bibliography{REF}

\begin{thebibliography}{56}
\providecommand{\natexlab}[1]{#1}

\bibitem[{ACO()}]{ACOSharmaLiteratureDatasets}

\newblock {{ACOSharma}}/literature {$\cdot$} {{Datasets}} at {{Hugging Face}}.
\newblock https://huggingface.co/datasets/ACOSharma/literature.

\bibitem[{Ale()}]{AlekseyKorshukFairytalebooksDatasets}

\newblock {{AlekseyKorshuk}}/fairy-tale-books {$\cdot$} {{Datasets}} at {{Hugging Face}}.
\newblock https://huggingface.co/datasets/AlekseyKorshuk/fairy-tale-books.

\bibitem[{Ano()}]{AnonymizedRepositoryAnonymousb}

\newblock Anonymized {{Repository}} - {{Anonymous GitHub}}.
\newblock https://anonymous.4open.science/r/Babel-3EB2/README.md.

\bibitem[{Twa()}]{Twang2218ChineselawregulationsDatasets}

\newblock Twang2218/chinese-law-and-regulations {$\cdot$} {{Datasets}} at {{Hugging Face}}.
\newblock https://huggingface.co/datasets/twang2218/chinese-law-and-regulations.

\bibitem[{chi(2020)}]{chinesepeoplesurvey}
 2020.
\newblock \href {https://www.chinahighlights.com/travelguide/english-levels-in-china.htm#/} {English levels in china: Quality of spoken english, signage, etc.}

\bibitem[{Sen(2024)}]{SentencetransformersAllMiniLML6v2Hugging2024}
 2024.
\newblock Sentence-transformers/all-{{MiniLM-L6-v2}} {$\cdot$} {{Hugging Face}}.
\newblock https://huggingface.co/sentence-transformers/all-MiniLM-L6-v2.

\bibitem[{Cla(2025)}]{Claude37Sonnet}
 2025.
\newblock \href {https://www.anthropic.com/news/claude-3-7-sonnet} {Claude 3.7 {{Sonnet}} and {{Claude Code}}}.

\bibitem[{Hel(2025)}]{HelloGPT4o}
 2025.
\newblock \href {https://openai.com/index/hello-gpt-4o/} {Hello {{GPT-4o}}}.

\bibitem[{Association(2019)}]{associationPublicationManualAmerican2019}
American~Psychological Association. 2019.
\newblock \href {https://faculty.tamuc.edu/jdavis/tmgt/599/242/TMGT599-242-Syllabus.pdf} {Publication manual of the american psychological association,(2020)}.
\newblock 428.

\bibitem[{Bagdasaryan and Shmatikov(2022)}]{bagdasaryan2022spinning}
Eugene Bagdasaryan and Vitaly Shmatikov. 2022.
\newblock Spinning language models: Risks of propaganda-as-a-service and countermeasures.
\newblock In \emph{2022 IEEE Symposium on Security and Privacy (SP)}, pages 1532--1532.

\bibitem[{Carlini et~al.(2021)Carlini, Tramer, Wallace, Jagielski, Herbert-Voss, Lee, Roberts, Brown, Song, Erlingsson et~al.}]{carlini2021extracting}
Nicholas Carlini, Florian Tramer, Eric Wallace, Matthew Jagielski, Ariel Herbert-Voss, Katherine Lee, Adam Roberts, Tom Brown, Dawn Song, Ulnar Erlingsson, et~al. 2021.
\newblock Extracting training data from large language models.
\newblock In \emph{USENIX Security Symposium}, pages 2633--2650.

\bibitem[{Chandrasekaran and Mago(2022)}]{chandrasekaranEvolutionSemanticSimilarity2022}
Dhivya Chandrasekaran and Vijay Mago. 2022.
\newblock \href {https://doi.org/10.1145/3440755} {Evolution of {{Semantic Similarity}} -- {{A Survey}}}.
\newblock \emph{ACM Computing Surveys}, 54(2):1--37.

\bibitem[{Chen(2024)}]{chenLMStyleBenchmarkEvaluating2024}
Jianlin Chen. 2024.
\newblock \href {https://doi.org/10.48550/arXiv.2403.08943} {{{LMStyle Benchmark}}: {{Evaluating Text Style Transfer}} for {{Chatbots}}}.
\newblock \emph{Preprint}, arXiv:2403.08943.

\bibitem[{Chen et~al.(2022)Chen, Liu, Haque, Song, and Yang}]{chen2022nmtsloth}
Simin Chen, Cong Liu, Mirazul Haque, Zihe Song, and Wei Yang. 2022.
\newblock Nmtsloth: understanding and testing efficiency degradation of neural machine translation systems.
\newblock In \emph{Proceedings of the 30th ACM Joint European Software Engineering Conference and Symposium on the Foundations of Software Engineering}, pages 1148--1160.

\bibitem[{Conneau et~al.(2020)Conneau, Khandelwal, Goyal, Chaudhary, Wenzek, Guzmán, Grave, Ott, Zettlemoyer, and Stoyanov}]{conneauUnsupervisedCrosslingualRepresentation2020}
Alexis Conneau, Kartikay Khandelwal, Naman Goyal, Vishrav Chaudhary, Guillaume Wenzek, Francisco Guzmán, Edouard Grave, Myle Ott, Luke Zettlemoyer, and Veselin Stoyanov. 2020.
\newblock \href {https://doi.org/10.48550/arXiv.1911.02116} {Unsupervised {{Cross-lingual Representation Learning}} at {{Scale}}}.
\newblock \emph{Preprint}, arXiv:1911.02116.

\bibitem[{Cui et~al.(2016)Cui, Liu, Chen, Wang, and Hu}]{cui-etal-2016-consensus}
Yiming Cui, Ting Liu, Zhipeng Chen, Shijin Wang, and Guoping Hu. 2016.
\newblock Consensus attention-based neural networks for chinese reading comprehension.
\newblock In \emph{Proceedings of COLING 2016, the 26th International Conference on Computational Linguistics: Technical Papers}, pages 1777--1786, Osaka, Japan.

\bibitem[{Dai et~al.(2019)Dai, Liang, Qiu, and Huang}]{dai2019style}
Ning Dai, Jianze Liang, Xipeng Qiu, and Xuanjing Huang. 2019.
\newblock Style transformer: Unpaired text style transfer without disentangled latent representation.
\newblock \emph{arXiv preprint arXiv:1905.05621}.

\bibitem[{Dementieva et~al.(2023)Dementieva, Moskovskiy, Dale, and Panchenko}]{dementievaExploringMethodsCrosslingual2023}
Daryna Dementieva, Daniil Moskovskiy, David Dale, and Alexander Panchenko. 2023.
\newblock \href {https://doi.org/10.18653/v1/2023.ijcnlp-main.70} {Exploring {{Methods}} for {{Cross-lingual Text Style Transfer}}: {{The Case}} of {{Text Detoxification}}}.
\newblock In \emph{Proceedings of the 13th {{International Joint Conference}} on {{Natural Language Processing}} and the 3rd {{Conference}} of the {{Asia-Pacific Chapter}} of the {{Association}} for {{Computational Linguistics}} ({{Volume}} 1: {{Long Papers}})}, pages 1083--1101. Association for Computational Linguistics.

\bibitem[{Devlin et~al.(2018)Devlin, Chang, Lee, and Toutanova}]{devlinBertPretrainingDeep2018}
Jacob Devlin, Ming-Wei Chang, Kenton Lee, and Kristina Toutanova. 2018.
\newblock \href {https://arxiv.org/abs/1810.04805} {Bert: {{Pre-training}} of deep bidirectional transformers for language understanding}.

\bibitem[{Fishman(2020)}]{fishman2020speaks}
Joshua~A Fishman. 2020.
\newblock Who speaks what language to whom and when?
\newblock In \emph{The bilingualism reader}, pages 55--70. Routledge.

\bibitem[{Fu et~al.(2018)Fu, Tan, Peng, Zhao, and Yan}]{fu2018style}
Zhenxin Fu, Xiaoye Tan, Nanyun Peng, Dongyan Zhao, and Rui Yan. 2018.
\newblock Style transfer in text: Exploration and evaluation.
\newblock In \emph{Proceedings of the AAAI conference on artificial intelligence}, volume~32.

\bibitem[{Gupta et~al.(2020)Gupta, He, Meister, and Su}]{gupta2020machine}
Shashij Gupta, Pinjia He, Clara Meister, and Zhendong Su. 2020.
\newblock Machine translation testing via pathological invariance.
\newblock In \emph{Proceedings of the 28th ACM Joint Meeting on European Software Engineering Conference and Symposium on the Foundations of Software Engineering}, pages 863--875.

\bibitem[{Han et~al.(2023)Han, Kumar, and Tsvetkov}]{hanSSDLMSemiautoregressiveSimplexbased2023}
Xiaochuang Han, Sachin Kumar, and Yulia Tsvetkov. 2023.
\newblock \href {https://doi.org/10.48550/arXiv.2210.17432} {{{SSD-LM}}: {{Semi-autoregressive Simplex-based Diffusion Language Model}} for {{Text Generation}} and {{Modular Control}}}.
\newblock \emph{Preprint}, arxiv:2210.17432.

\bibitem[{He et~al.(2020)He, Meister, and Su}]{he2020structure}
Pinjia He, Clara Meister, and Zhendong Su. 2020.
\newblock Structure-invariant testing for machine translation.
\newblock In \emph{Proceedings of the ACM/IEEE 42nd International Conference on Software Engineering}, pages 961--973.

\bibitem[{He et~al.(2021)He, Meister, and Su}]{he2021testing}
Pinjia He, Clara Meister, and Zhendong Su. 2021.
\newblock Testing machine translation via referential transparency.
\newblock In \emph{2021 IEEE/ACM 43rd International Conference on Software Engineering (ICSE)}, pages 410--422. IEEE.

\bibitem[{Heigold et~al.(2017)Heigold, Neumann, and van Genabith}]{heigold2017robust}
Georg Heigold, G{\"u}nter Neumann, and Josef van Genabith. 2017.
\newblock How robust are character-based word embeddings in tagging and mt against wrod scramlbing or randdm nouse?
\newblock \emph{arXiv preprint arXiv:1704.04441}.

\bibitem[{Horvitz et~al.(2024)Horvitz, Patel, Callison-Burch, Yu, and McKeown}]{horvitzParaGuideGuidedDiffusion2024}
Zachary Horvitz, Ajay Patel, Chris Callison-Burch, Zhou Yu, and Kathleen McKeown. 2024.
\newblock \href {https://doi.org/10.48550/arXiv.2308.15459} {{{ParaGuide}}: {{Guided Diffusion Paraphrasers}} for {{Plug-and-Play Textual Style Transfer}}}.
\newblock \emph{Preprint}, arxiv:2308.15459.

\bibitem[{Hovy et~al.(2020)Hovy, Bianchi, and Fornaciari}]{hovyYouSoundJust2020}
Dirk Hovy, Federico Bianchi, and Tommaso Fornaciari. 2020.
\newblock \href {https://doi.org/10.18653/v1/2020.acl-main.154} {“{{You Sound Just Like Your Father}}” {{Commercial Machine Translation Systems Include Stylistic Biases}}}.
\newblock In \emph{Proceedings of the 58th {{Annual Meeting}} of the {{Association}} for {{Computational Linguistics}}}, pages 1686--1690. Association for Computational Linguistics.

\bibitem[{Hu et~al.(2017)Hu, Yang, Liang, Salakhutdinov, and Xing}]{hu2017toward}
Zhiting Hu, Zichao Yang, Xiaodan Liang, Ruslan Salakhutdinov, and Eric~P Xing. 2017.
\newblock Toward controlled generation of text.
\newblock In \emph{International conference on machine learning}, pages 1587--1596. PMLR.

\bibitem[{Ji et~al.(2021)Ji, Feng, Liu, Zhao, and Xu}]{ji2021automated}
Pin Ji, Yang Feng, Jia Liu, Zhihong Zhao, and Baowen Xu. 2021.
\newblock Automated testing for machine translation via constituency invariance.
\newblock In \emph{2021 36th IEEE/ACM International Conference on Automated Software Engineering (ASE)}, pages 468--479. IEEE.

\bibitem[{Jin et~al.(2022)Jin, Jin, Hu, Vechtomova, and Mihalcea}]{jinDeepLearningText2022}
Di~Jin, Zhijing Jin, Zhiting Hu, Olga Vechtomova, and Rada Mihalcea. 2022.
\newblock \href {https://doi.org/10.1162/coli_a_00426} {Deep {{Learning}} for {{Text Style Transfer}}: {{A Survey}}}.
\newblock 48(1):155--205.

\bibitem[{Jin et~al.(2019)Jin, Dhingra, Liu, Cohen, and Lu}]{jin2019pubmedqa}
Qiao Jin, Bhuwan Dhingra, Zhengping Liu, William Cohen, and Xinghua Lu. 2019.
\newblock Pubmedqa: A dataset for biomedical research question answering.
\newblock In \emph{Proceedings of the 2019 Conference on Empirical Methods in Natural Language Processing and the 9th International Joint Conference on Natural Language Processing (EMNLP-IJCNLP)}, pages 2567--2577.

\bibitem[{Kang and Hovy(2021)}]{kangStyleNOTSingle2021}
Dongyeop Kang and Eduard Hovy. 2021.
\newblock \href {https://doi.org/10.18653/v1/2021.acl-long.185} {Style is {{NOT}} a single variable: {{Case Studies}} for {{Cross-Stylistic Language Understanding}}}.
\newblock In \emph{Proceedings of the 59th {{Annual Meeting}} of the {{Association}} for {{Computational Linguistics}} and the 11th {{International Joint Conference}} on {{Natural Language Processing}} ({{Volume}} 1: {{Long Papers}})}, pages 2376--2387. Association for Computational Linguistics.

\bibitem[{Li et~al.(2022{\natexlab{a}})Li, Bhambhoria, and Zhu}]{li-etal-2022-parameter}
Jonathan Li, Rohan Bhambhoria, and Xiaodan Zhu. 2022{\natexlab{a}}.
\newblock \href {https://aclanthology.org/2022.nllp-1.10} {Parameter-efficient legal domain adaptation}.
\newblock In \emph{Proceedings of the Natural Legal Language Processing Workshop 2022}, pages 119--129, Abu Dhabi, United Arab Emirates (Hybrid). Association for Computational Linguistics.

\bibitem[{Li et~al.(2022{\natexlab{b}})Li, Thickstun, Gulrajani, Liang, and Hashimoto}]{li2022diffusion}
Xiang Li, John Thickstun, Ishaan Gulrajani, Percy~S Liang, and Tatsunori~B Hashimoto. 2022{\natexlab{b}}.
\newblock Diffusion-lm improves controllable text generation.
\newblock \emph{Advances in Neural Information Processing Systems}, 35:4328--4343.

\bibitem[{Liu et~al.(2024)Liu, Qin, Ye, Mou, He, and Wang}]{liuAdaptivePromptRouting2024}
Qingyi Liu, Jinghui Qin, Wenxuan Ye, Hao Mou, Yuxuan He, and Keze Wang. 2024.
\newblock \href {https://doi.org/10.1609/aaai.v38i17.29832} {Adaptive prompt routing for arbitrary text style transfer with pre-trained language models}.
\newblock In \emph{Proceedings of the {{Thirty-Eighth AAAI Conference}} on {{Artificial Intelligence}} and {{Thirty-Sixth Conference}} on {{Innovative Applications}} of {{Artificial Intelligence}} and {{Fourteenth Symposium}} on {{Educational Advances}} in {{Artificial Intelligence}}}, volume~38 of \emph{{{AAAI}}'24/{{IAAI}}'24/{{EAAI}}'24}, pages 18689--18697. AAAI Press.

\bibitem[{Luo et~al.(2023)Luo, Han, Mou, and Firdaus}]{luoPromptBasedEditingText2023}
Guoqing Luo, Yu~Han, Lili Mou, and Mauajama Firdaus. 2023.
\newblock \href {https://doi.org/10.18653/v1/2023.findings-emnlp.381} {Prompt-{{Based Editing}} for {{Text Style Transfer}}}.
\newblock In \emph{Findings of the {{Association}} for {{Computational Linguistics}}: {{EMNLP}} 2023}, pages 5740--5750. Association for Computational Linguistics.

\bibitem[{Mai et~al.(2023)Mai, Jiang, and Deng}]{maiPrefixTuningBasedUnsupervised2023}
Huiyu Mai, Wenhao Jiang, and Zhi-Hong Deng. 2023.
\newblock \href {https://doi.org/10.18653/v1/2023.findings-emnlp.990} {Prefix-{{Tuning Based Unsupervised Text Style Transfer}}}.
\newblock In \emph{Findings of the {{Association}} for {{Computational Linguistics}}: {{EMNLP}} 2023}, pages 14847--14856. Association for Computational Linguistics.

\bibitem[{Merity et~al.(2016)Merity, Xiong, Bradbury, and Socher}]{merity2016pointer}
Stephen Merity, Caiming Xiong, James Bradbury, and Richard Socher. 2016.
\newblock \href {https://arxiv.org/abs/1609.07843} {Pointer sentinel mixture models}.
\newblock \emph{Preprint}, arXiv:1609.07843.

\bibitem[{Mukherjee et~al.(2023)Mukherjee, Bansal, Kr.~Ojha, P.~McCrae, and Dusek}]{mukherjeeTextDetoxificationStyle2023}
Sourabrata Mukherjee, Akanksha Bansal, Atul Kr.~Ojha, John P.~McCrae, and Ondrej Dusek. 2023.
\newblock \href {https://aclanthology.org/2023.icon-1.13/} {Text {{Detoxification}} as {{Style Transfer}} in {{English}} and {{Hindi}}}.
\newblock In \emph{Proceedings of the 20th {{International Conference}} on {{Natural Language Processing}} ({{ICON}})}, pages 133--144. NLP Association of India (NLPAI).

\bibitem[{Pan et~al.(2024)Pan, Lan, Li, and Qian}]{panUnsupervisedTextStyle2024}
Lei Pan, Yunshi Lan, Yang Li, and Weining Qian. 2024.
\newblock \href {https://doi.org/10.48550/arXiv.2402.13647} {Unsupervised {{Text Style Transfer}} via {{LLMs}} and {{Attention Masking}} with {{Multi-way Interactions}}}.
\newblock \emph{Preprint}, arXiv:2402.13647.

\bibitem[{Pesu et~al.(2018)Pesu, Zhou, Zhen, and Towey}]{pesu2018monte}
Daniel Pesu, Zhi~Quan Zhou, Jingfeng Zhen, and Dave Towey. 2018.
\newblock A monte carlo method for metamorphic testing of machine translation services.
\newblock In \emph{Proceedings of the 3rd International Workshop on Metamorphic Testing}, pages 38--45.

\bibitem[{Shen et~al.(2017)Shen, Lei, Barzilay, and Jaakkola}]{shen2017style}
Tianxiao Shen, Tao Lei, Regina Barzilay, and Tommi Jaakkola. 2017.
\newblock Style transfer from non-parallel text by cross-alignment.
\newblock \emph{Advances in neural information processing systems}, 30.

\bibitem[{Sheng et~al.(2021)Sheng, Chang, Natarajan, and Peng}]{sheng2021societal}
Emily Sheng, Kai-Wei Chang, Prem Natarajan, and Nanyun Peng. 2021.
\newblock Societal biases in language generation: Progress and challenges.
\newblock In \emph{Proceedings of the 59th Annual Meeting of the Association for Computational Linguistics}, pages 4275--4293.

\bibitem[{Sun et~al.(2024)Sun, Chen, Zhang, and Hao}]{sun2024fairness}
Zeyu Sun, Zhenpeng Chen, Jie Zhang, and Dan Hao. 2024.
\newblock Fairness testing of machine translation systems.
\newblock \emph{ACM Transactions on Software Engineering and Methodology}.

\bibitem[{Sun et~al.(2020)Sun, Zhang, Harman, Papadakis, and Zhang}]{sun2020automatic}
Zeyu Sun, Jie~M Zhang, Mark Harman, Mike Papadakis, and Lu~Zhang. 2020.
\newblock Automatic testing and improvement of machine translation.
\newblock In \emph{Proceedings of the ACM/IEEE 42nd international conference on software engineering}, pages 974--985.

\bibitem[{Sun et~al.(2022)Sun, Zhang, Xiong, Harman, Papadakis, and Zhang}]{sun2022improving}
Zeyu Sun, Jie~M Zhang, Yingfei Xiong, Mark Harman, Mike Papadakis, and Lu~Zhang. 2022.
\newblock Improving machine translation systems via isotopic replacement.
\newblock In \emph{Proceedings of the 44th international conference on software engineering}, pages 1181--1192.

\bibitem[{Tiedemann and Thottingal(2020)}]{TiedemannThottingal}
J{\"o}rg Tiedemann and Santhosh Thottingal. 2020.
\newblock {OPUS-MT} — {B}uilding open translation services for the {W}orld.
\newblock In \emph{Proceedings of the 22nd Annual Conferenec of the European Association for Machine Translation (EAMT)}, Lisbon, Portugal.

\bibitem[{Weidinger et~al.(2022)Weidinger, Uesato, Rauh, Griffin, Huang, Mellor, Glaese, Cheng, Balle, Kasirzadeh et~al.}]{weidinger2022taxonomy}
Laura Weidinger, Jonathan Uesato, Maribeth Rauh, Conor Griffin, Po-Sen Huang, John Mellor, Amelia Glaese, Myra Cheng, Borja Balle, Atoosa Kasirzadeh, et~al. 2022.
\newblock Taxonomy of risks posed by language models.
\newblock \emph{2022 ACM Conference on Fairness, Accountability, and Transparency}, pages 214--229.

\bibitem[{Xu(2019)}]{bright_xu_2019_3402023}
Bright Xu. 2019.
\newblock \href {https://doi.org/10.5281/zenodo.3402023} {Nlp chinese corpus: Large scale chinese corpus for nlp}.

\bibitem[{Xu et~al.(2017)Xu, Wen, Sun, and Su}]{dnerre}
Jingjing Xu, Ji~Wen, Xu~Sun, and Qi~Su. 2017.
\newblock A discourse-level named entity recognition and relation extraction dataset for chinese literature text.
\newblock volume abs/1711.07010.

\bibitem[{Zhang et~al.(2024{\natexlab{a}})Zhang, Cai, Li, Wu, Hou, and Abdul-Mageed}]{zhangDistillingTextStyle2024}
Chiyu Zhang, Honglong Cai, Yuezhang Li, Yuexin Wu, Le~Hou, and Muhammad Abdul-Mageed. 2024{\natexlab{a}}.
\newblock \href {https://doi.org/10.18653/v1/2024.naacl-srw.21} {Distilling {{Text Style Transfer With Self-Explanation From LLMs}}}.
\newblock In \emph{Proceedings of the 2024 {{Conference}} of the {{North American Chapter}} of the {{Association}} for {{Computational Linguistics}}: {{Human Language Technologies}} ({{Volume}} 4: {{Student Research Workshop}})}, pages 200--211. Association for Computational Linguistics.

\bibitem[{Zhang et~al.(2020)Zhang, Zhao, Saleh, and Liu}]{zhang2020pegasus}
Jingqing Zhang, Yao Zhao, Mohammad Saleh, and Peter Liu. 2020.
\newblock Pegasus: Pre-training with extracted gap-sentences for abstractive summarization.
\newblock In \emph{International conference on machine learning}, pages 11328--11339. PMLR.

\bibitem[{Zhang et~al.(2022)Zhang, Chen, Bi, Liang, Li, Shang, Yin, Tan, Xu, Huang, Si, Ni, Xie, Sui, Chang, Zong, Yuan, Li, Yan, Zan, Zhang, Tang, and Chen}]{zhang-etal-2022-cblue}
Ningyu Zhang, Mosha Chen, Zhen Bi, Xiaozhuan Liang, Lei Li, Xin Shang, Kangping Yin, Chuanqi Tan, Jian Xu, Fei Huang, Luo Si, Yuan Ni, Guotong Xie, Zhifang Sui, Baobao Chang, Hui Zong, Zheng Yuan, Linfeng Li, Jun Yan, Hongying Zan, Kunli Zhang, Buzhou Tang, and Qingcai Chen. 2022.
\newblock \href {https://doi.org/10.18653/v1/2022.acl-long.544} {{CBLUE}: A {C}hinese biomedical language understanding evaluation benchmark}.
\newblock In \emph{Proceedings of the 60th Annual Meeting of the Association for Computational Linguistics (Volume 1: Long Papers)}, pages 7888--7915, Dublin, Ireland. Association for Computational Linguistics.

\bibitem[{Zhang et~al.(2024{\natexlab{b}})Zhang, Zhai, Fang, Liu, Sun, Hu, and Wang}]{zhang2024machine}
Quanjun Zhang, Juan Zhai, Chunrong Fang, Jiawei Liu, Weisong Sun, Haichuan Hu, and Qingyu Wang. 2024{\natexlab{b}}.
\newblock Machine translation testing via syntactic tree pruning.
\newblock \emph{ACM Transactions on Software Engineering and Methodology}.

\bibitem[{Zhang et~al.(2018)Zhang, Ren, Liu, Wang, Chen, Li, Zhou, and Chen}]{zhang2018style}
Zhirui Zhang, Shuo Ren, Shujie Liu, Jianyong Wang, Peng Chen, Mu~Li, Ming Zhou, and Enhong Chen. 2018.
\newblock Style transfer as unsupervised machine translation.
\newblock \emph{arXiv preprint arXiv:1808.07894}.

\end{thebibliography}

\appendix

\section{\sys-Corpus}\label{sec:corpus}

\begin{table*}[]
    \caption{Datasets being extracted to \sys-Corpus.}\label{tab:ev}
    \centering
    \scalebox{0.8}{
        \begin{tabular}{lll}
        \toprule 
        Domain                    & English                     & Chinese                    \\ \midrule
        Law                       & Law Stack Exchange~\cite{li-etal-2022-parameter}          & ChineseLaw and Regulations~\cite{Twang2218ChineselawregulationsDatasets}   \\
        Literature                & Classic Literature in ASCII~\cite{ACOSharmaLiteratureDatasets}   & Chinese Literature~\cite{dnerre}           \\
        Wikipedia                 & wikitext~\cite{merity2016pointer}                      & wiki2019zh~\cite{bright_xu_2019_3402023}                   \\
        Medicine                  & PubmedQA~\cite{jin2019pubmedqa}                      & CBLUE~\cite{zhang-etal-2022-cblue}                        \\
        Early Childhood Education & Fairy Tale Books~\cite{AlekseyKorshukFairytalebooksDatasets}              & CFT~\cite{cui-etal-2016-consensus}                          \\ \bottomrule
        \end{tabular}
    }
\end{table*}

The selected data has distinct stylistic features and maintains the stylistic correspondence between Chinese and English, making it suitable for evaluating our method.
Our preprocessing steps include metadata removal, formatting standardization, tokenization, sentence segmentation, and ensuring that each data point contained at least one complete sentence.

\noindent
\textbf{Ethical and privacy considerations}
In the compilation of our dataset, we have been vigilant in addressing ethical and privacy concerns to ensure that the data utilized does not infringe upon individual rights or breach any legal regulations. 
Our dataset is derived from publicly available resources, and we have taken the following measures to uphold ethical standards and protect privacy:
\begin{itemize}
    \item All personal identifiers have been removed from the data to prevent the identification of individuals. This process includes the removal of names, addresses, and any other unique identifiers that could be linked to specific individuals.
    \item Where applicable, we have obtained necessary permissions and consents from the original data providers or authors of the texts to use the data for research purposes.
    \item The dataset has been reviewed by an independent ethics committee to ensure that it meets the ethical standards required for academic research.
\end{itemize}

\section{Detailed setup of \sys}\label{sec:setupapp}

\subsection{Software and Hardware}
We conduct our experiments on a server with 64 cores Intel Xeon 2.90GHz CPU, 256 GB RAM, and 4 NVIDIA 3090 GPUs running the Ubuntu 20.04 operating system.

\subsection{Translation Systems}

\noindent
\textbf{Google Translate.}
Google Translate is a multilingual neural machine translation system developed by Google.
It supports over 100 languages, has a vast user base with over 500 million users and translates more than 100 billion words daily.

\noindent
\textbf{Baidu Translate.}
Baidu Translate is optimized for translations between Chinese and other languages, leveraging Baidu's AI and big data technologies. 
It serves millions of users in China, providing text, voice, and image translation services.

\noindent
\textbf{Youdao Translate.}
Youdao Translate, developed by NetEase, integrates rich dictionary resources and NMT technology for accurate translations, particularly beneficial for educational purposes.
It is widely used by students and educators in China, with millions of active users.

\noindent
\textbf{Opus-MT.}
Opus-MT is an open-source neural machine translation model based on the Transformers architecture, supported by the Marian NMT toolkit. 
It is popular among researchers and developers for its flexibility and customization options, and there were 1.55M downloads on huggingface last month. 

\subsection{Human Evaluation}\label{sec:human}
We engaged ten annotators for the evaluation process: three native Chinese speakers~(live in China) proficient in English, and two native English speakers proficient in Chinese~(live in Singapore). 
All annotators hold advanced degrees, with at least an undergraduate qualification, and have professional expertise in linguistics, translation studies, or literature. 
These qualifications ensured that the evaluation was conducted by individuals with the necessary expertise to assess translation quality accurately. 
Annotators were recruited based on their professional backgrounds and were not compensated for their participation, as the study was conducted in-house with experts who volunteered due to their academic and professional interests. 
Consent was obtained from all annotators before their involvement, with instructions provided that explained how their evaluation data~(ratings and feedback) would be used for research purposes. 
The instructions also outlined the evaluation criteria and expectations for the task, ensuring full transparency. 
No risks to participants were identified, and participation was voluntary. 
The data collection protocol was exempt from formal ethics review, as it involved professional annotators in a controlled research setting, and all procedures adhered to ethical standards for transparency and informed consent.

\section{Style Detector Performance Comparison}\label{sec:detectorcomparison}

We conducted experiments comparing different approaches for style detection: (1) using separate language-specific BERT models for source and target languages, (2) using separate language-specific XLM-R models, and (3) using a single cross-lingual XLM-R model for both languages.
For the separate BERT approach, we fine-tuned individual BERT-base models for Chinese and English. 
For the separate XLM-R approach, we fine-tuned separate XLM-R models for each language. 
For the cross-lingual approach, we fine-tuned a single XLM-R model to handle both languages simultaneously.
\autoref{tab:style-detector} presents the style classification accuracy results across our five domains for all three approaches.

\begin{table}[tb]
\caption{Style classification accuracy comparison across model configurations.}\label{tab:style-detector}
\centering
\scalebox{0.4}{
    \begin{tabular}{crrrr}\toprule
        Domain     & Separate BERT Models & Separate XLM-R Models & Cross-lingual XLM-R \\ \midrule
        Law        & 88.4\%               & 82.1\%                & 77.2\%              \\
        Literature & 91.6\%               & 83.8\%                & 78.5\%              \\
        Wikipedia  & 89.3\%               & 84.0\%                & 79.1\%              \\
        Medicine   & 93.2\%               & 85.7\%                & 79.8\%              \\
        Education  & 90.7\%               & 82.6\%                & 78.3\%              \\ \midrule
        Average    & 90.6\%               & 83.6\%                & 78.6\%              \\ \bottomrule
        \end{tabular}
}
\end{table}

The results demonstrate that separate BERT models consistently outperform the cross-lingual XLM-R model by an average of 12.0\%. 
Even separate XLM-R models outperform the cross-lingual XLM-R by 5.0\% on average. 
This suggests that while cross-lingual models offer convenience by handling multiple languages within a single model, they sacrifice accuracy in capturing language-specific stylistic nuances.
This finding informed our decision to use separate language-specific BERT models for style detection in Babel, prioritizing accuracy in stylistic analysis over the convenience of a single cross-lingual model.

\section{Technical Details}\label{sec:tech}

\begin{algorithm}[t]
    \small
    \caption{Style Applicator of \sys}
    \label{alg:overview}
    \footnotesize
    \begin{algorithmic}[1]
        
        

        \item[]
        \Require \(\mathbf{r}\): initial translated texts
        \Require \(\mathbf{y}\): user-supplied sample texts
        \Require \(D_{\theta^*}(\cdot)\): trained diffusion model 
        \Require \(T\): number of total diffusion step
        \Require \(\lambda\): number of total diffusion step
        \Ensure \(\mathbf{r}^*\): optimized translated texts
        \Procedure{ApplicateStyle}{}
        \State $\mathbf{x}_T \gets SampleFrom(\mathcal{N}(0,\mathbf{I}))$ \Comment{\textit{ Sample a random gaussian noise}}
        \For{$t \gets T \  to \ 1 $}
        \State $\mathbf{\hat l}_t \gets D_{\theta^*}(\mathbf{x}_t,t,\mathbf{r})$
        \State $\mathbf{\hat r}_t \gets SampleFrom(Top\text{-}p(Softmax(\mathbf{\hat l}_t)))$
        \State $J \gets GetSimilarity(\mathbf{\hat r}_t,\mathbf{y})$
        \State $\mathbf{\hat l}_t^* \gets \mathbf{\hat l}_t-\lambda GetGradient(J)$
        \State $\mathbf{\hat r}_t^* \gets SampleFrom(Top\text{-}p(Softmax(\mathbf{\hat l}_t^*)))$
        \State $\mathbf{x}_{t-1} \gets ForwardDiffusion(\mathbf{\hat r}_0^*)$
        \EndFor
        \State $\mathbf{\hat l}_0 \gets D_{\theta^*}(\mathbf{x}_0,0,\mathbf{r})$
        \State $\mathbf{\hat r}_0 \gets SampleFrom(Top\text{-}p(Softmax(\mathbf{\hat l}_0)))$
        \State $J \gets GetSimilarity(\mathbf{\hat r}_0,\mathbf{y})$
        \State $\mathbf{\hat l}_0^* \gets \mathbf{\hat l}_t-\lambda GetGradient(J)$
        \State $\mathbf{\hat r}_0^* \gets SampleFrom(Top\text{-}p(Softmax(\mathbf{\hat l}_0^*)))$
        
        \Return $\mathbf{\hat r}_0^*$ 
        \EndProcedure  
    \end{algorithmic}

\end{algorithm}

\subsection{Training Process Formulation}\label{sec:trainingproc}





To imitate the style loss observed in translation, we use a paraphrase model \(P(\cdot)\), such as PEGASUS~\cite{zhang2020pegasus}, to generate paraphrases \(\mathbf{p}\) of the input text \(\mathbf{r}\). 
Formally, we have:
\begin{equation}\label{eq:paraphrase1}
    \mathbf{p} = P(\mathbf{r})
\end{equation}
These paraphrases retain the original meaning but have reduced stylistic elements, simulating the effect of translation where the core content remains intact, but the style may be neutralized.
This step is crucial for preparing the model to neutralize and extract the stylistic essence of sentences while preserving their semantic content.

The diffusion is performed in the embedding space, where the text is represented in a numerical format that captures its meaning. 
Operating in the embedding space helps maintain the semantic integrity of the sentences. 
Formally, we have:
\begin{equation}\label{eq:sample1}
    \mathbf{x}_t = \sqrt{\beta_t}E(\mathbf{r}) + \sqrt{(1-\beta_t)}\bm{\epsilon}_t \quad\quad \bm{\epsilon_t} \sim \mathcal N(0,\mathbf{I})
\end{equation}
where \(E(\cdot)\) is an embedding model.
In terms of the schedule of noise, we follow the paradigm of \citet{horvitzParaGuideGuidedDiffusion2024}, that is:
\begin{equation}\label{eq:schedule1}
    \beta_t = \sqrt{\frac{T-t}{T}}
\end{equation}
This schedule decreases to zero at a significantly slower rate compared to the cosine and square root schedules, thus preserving information more effectively. 
For NLP tasks, this feature is crucial as it helps to maintain the semantic information of the original text.

After completing these preparations above, we perform the training process on the paraphrased text, as mentioned in \autoref{sec:diffusion}.
Formally, we train the model \(D_\theta(\cdot)\) by minimizing the cross entropy between the posterior distribution of the model at each diffusion time step and the actual embeddings:
\begin{equation}\label{eq:training1}
    \mathcal{L}(\theta) = \mathcal{E}\left[\log p_\theta(\mathbf{r}|D_\theta(\mathbf{x}_t,t,\mathbf{p}))\right] 
\end{equation}
where \(\mathcal{L}(\cdot)\) is the loss function, \(\mathcal{E}(\cdot)\) represents the cross entropy function, \(\mathbf{r}\) is the original text, \(t\) represents the time step, and \(\mathbf{p}\) represents the paraphrase.
By making small adjustments at each step, the model turns data from the noisy state to the desired state. 
During this process, the model learns to preserve semantic content and reconstruct the original embeddings as closely as possible. 

\subsection{Inference Process Formulation}\label{sec:inferproc}

After completing the training, the diffusion model \(D_{\theta^*}(\cdot)\) can then be used to attach attributes to the text, a process we refer to as \textit{inference process}.

The inference process starts with sampling initial noisy data \(\mathbf{x}_T \sim \mathcal N(0,\mathbf{I})\) and iteratively removes the noise to construct the improved sentences. 
For each time step \(t\)~(\(t \in \left[T,1\right]\)), the style applicator estimates an optimized text:
\begin{equation}\label{eq:inference1}
    \mathbf{\hat r}_t \sim \text{top-p}(\text{softmax}((D_{\theta^*}(\mathbf{x}_t,t,\mathbf{r})))
\end{equation}
where \(\mathbf{r}\) represents initial translated texts output by translation system. 

The advantage of our style applicator is that the generated text can be gradient-guided based on user-supplied style samples, directing the output to a specific target style.
Given a set of user-supplied style samples \(\left[\mathbf{y_1},\cdots, \mathbf{y_n}\right]\) and a style embedding model \(E_s(\cdot)\), we can obtain the style guidance function
\begin{equation}\label{eq:guidance1}
    J = \frac{\sum_{i=1}^n d(E_s(\mathbf{\hat r}_t),E_s(\mathbf{y_i}))}{n}
\end{equation}
where \(d(\cdot,\cdot)\) represents cosine similarity.
So, we get the final style-guided textual inference equation:
\begin{equation}\label{eq:inference21}
    \mathbf{\hat r}_t^* \sim \text{top-p}(\text{softmax}((D_{\theta^*}(\mathbf{x}_t,t,\mathbf{r}))-\lambda\nabla J))
\end{equation}

After estimating \(\mathbf{\hat r}_t^*\),  we proceed backward in time to iteratively acquire states with proceeding time steps.
Similarly to the training process, the style applicator embeds these tokens using the word embedding model \(E(\cdot)\) and subsequently adds noise to generate the latent representation for the preceding diffusion time step:
\begin{equation}\label{eq:addnoise1}
    \mathbf{x}_{t-1} = \sqrt{\beta_{t-1}}E(\mathbf{\hat r}_t^*) + \sqrt{(1-\beta_{t-1})}\bm{\epsilon} \quad \bm{\epsilon} \sim \mathcal N(0,\mathbf{I})
\end{equation}
After iterating this process until \(t=0\), we eventually get the desired output \(\mathbf{\hat r}_0^*\).

\subsection{Model Training}

The two core models of \sys are configured as follows:

\noindent
\textbf{Style Detector.}
We start with the publicly available BERT-base-cased checkpoint\footnote{https://huggingface.co/google-bert/bert-base-cased} and BERT-base-chinese checkpoint\footnote{https://huggingface.co/google-bert/bert-base-chinese}, both equipped with a classification head.
Our model is trained for 200 steps on the \sys-Corpus~(train-test ratio is set to 8:2), with a batch size of 16 and a learning rate of 2e-5.
The style classification threshold \(h\) is set to 0.5, and we explore the impact of this parameter on style detection in~\autoref{sec:rq4}.

\noindent
\textbf{Style Applicator.}
We employ the publicly available SSDLM RoBERTa-large checkpoint~\cite{horvitzParaGuideGuidedDiffusion2024} and train our model for 200K steps on train set of \sys-Corpus, with a batch size of 128, total time steps of 800, and learning rate of 1e-5.
During inference, we use temperature \(\tau=0.3\) and guidance strength \(\lambda=1000\).
We investigate effects of these parameters in~\autoref{sec:rq4}.

\noindent
\textbf{User Customization.}
Users can customize the style they wish to address by providing samples of bilingual texts that exhibit the desired style. 
This process involves collecting a sufficient number of bilingual texts within the same style domain and fine-tuning them using a script we provide~\cite{AnonymizedRepositoryAnonymousb}.
For detailed information on the training cost, please refer to~\autoref{sec:rq3}.
It is important to note that, although this paper focuses on Chinese-English bilingual style repair due to resource constraints, \sys is theoretically applicable to any bilingual style text repair.
To adapt \sys for other language pairs, such as English-German, users need to provide English and German text samples of the target style and replace the base models in the style detector and style applicator.
Specifically, bert-base-chinese should be replaced with a German BERT model, such as BERT-base-german-cased\footnote{https://huggingface.co/google-bert/bert-base-german-cased}, and the SSDLM RoBERTa model should be substituted with a German large-language model, like xlm-roberta-german\footnote{https://huggingface.co/FacebookAI/xlm-roberta-large-finetuned-conll03-german}.

\subsection{Evaluation metrics}

\noindent
\textbf{Bias Ratio.}
We utilized the style detector to quantify the number of stylistic bias in the outputs of each translation system and to determine the proportion of bias relative to the total sample (refer to~\autoref{sec:rq1} for the detailed methodology). 
To ensure the validity of the style detector, we conducted a manual evaluation for confirmation (see~\autoref{sec:rq4}).

\noindent
\textbf{Style Score.}
To evaluate the overall stylistic bias of the translation system, we calculate the average style scores of all its outputs.
These style scores are derived from the confidence provided by the style detector.
Due to varying sentence lengths and stylistic distinctiveness across datasets, this score lacks absolute significance and is meaningful only when comparing different translation systems on the same dataset.

\noindent
\textbf{Semantic Textual Similarity.}
STS~(Semantic Textual Similarity)~\cite{chandrasekaranEvolutionSemanticSimilarity2022} is a criterion that assesses how similar two texts are in terms of meaning.
Since our focus lies in assessing the ability to repair translations without parallel texts, we calculate the STS score between the revised text and the initial translated text to gauge \sys's proficiency in preserving semantic integrity.
We use one of the most commonly used models for this task, \textit{all-MiniLM-L6-v2}~\cite{SentencetransformersAllMiniLML6v2Hugging2024}, for this assessment.

\section{Efficiency in Testing and Repairing stylistically inconsistent Bias}
\label{sec:rq3}
\begin{table}[]
    \caption{The time overhead of \sys. The values in the table are averaged over the entire dataset, in units of seconds.}\label{tab:rq3}
    \centering
    \scalebox{0.6}{
        \begin{tabular}{lrr}
            \toprule
            Translation System & Testing Cost & Repairing Cost \\ \midrule
            Google             & 1.81             & 4.14        \\
            Baidu              & 1.60             & 3.97        \\
            Youdao             & 1.74             & 3.69        \\
            Opus-MT               & 1.67             & 3.73        \\ \bottomrule
            \end{tabular}
    }
  \end{table}

\noindent
{\bf Experiment Design}:
To assess efficiency, we meticulously measure the time \sys expends during both the testing and repair phases for stylistic inconsistencies. For each translation system involved, we calculate the average duration required by \sys to complete a single cycle of stylistic bias detection and subsequent rectification. This comprehensive timing analysis enables us to determine the operational speed of \sys, ensuring it efficiently addresses style bias without significantly detracting from user experience, thereby maintaining seamless workflow integration.

\noindent
{\bf Results}:
The results are presented in~\autoref{tab:rq3}.
On average, testing a single original translation text with \sys requires only 1.7 seconds, while repairing a problematic text takes just 3.9 seconds. 
This demonstrates that \sys is efficient in both testing and repairing, improving translation style without significantly impacting user experience.

\noindent
{\bf Training cost}:
To estimate the computational cost for users adding a new style, we measured the time required to fine-tune a fifth domain style fix on a \sys that already supports four domains.
To mitigate the impact of individual datasets, we performed fine-tuning on each of the five domains separately and calculated the average time as the result.
Our experiments, conducted using the computational resources described in~\autoref{sec:setupapp}, indicate that fine-tuning a pair of datasets, each containing 1,000 samples, takes an average of 37,261 seconds~(approximately 10.5 hours).
It is important to note that the computational cost may vary substantially depending on the size of the datasets and the specific languages involved.

\section{Impact of Configurable Parameters}
\label{sec:rq4}

\begin{figure}
    \centering   
    
  \begin{subfigure}[figure1]{0.4\linewidth}  
        \centering
    \scalebox{0.7}{
    \includegraphics[height=3.5cm]{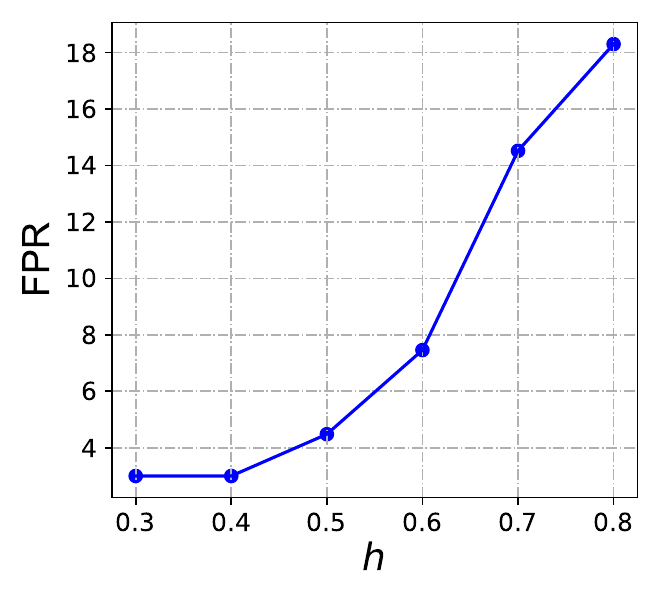} 
    }
      \caption{\(h\)-FPR}
      \label{fig:h-fpr}
  \end{subfigure}
  \begin{subfigure}[figure2]{0.4\linewidth}
        \centering
    \scalebox{0.7}{
    \includegraphics[height=3.5cm]{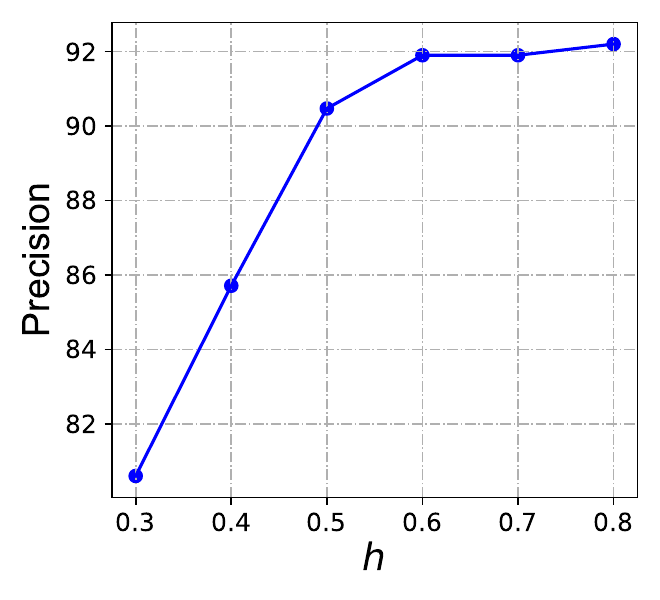}
    }
      \caption{\(h\)-precision}
      \label{fig:h-precision}
  \end{subfigure}
  
  \caption{Effect of \(h\) on the average performance of \sys's testing process.}
  \label{fig:h}
\end{figure}
\begin{figure*}
    \centering    
    \scalebox{0.7}{
    \subfloat[\(\tau\)-issues]
    {\includegraphics[width=0.23\textwidth,height=0.2\textwidth]{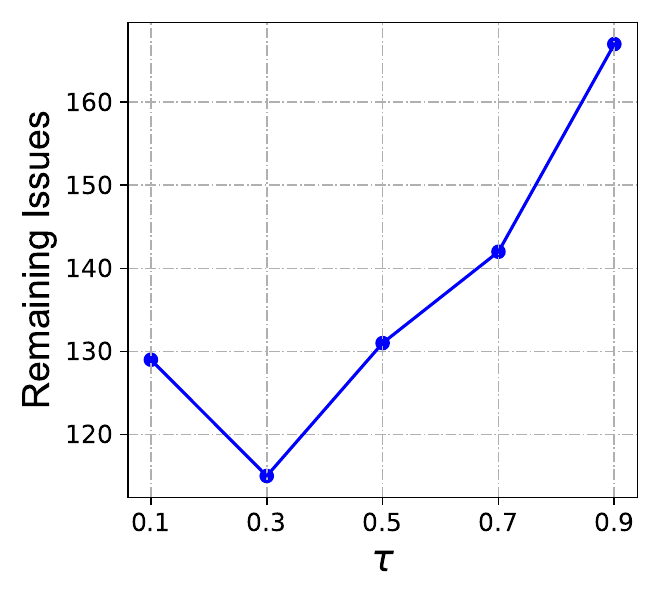}
    \label{fig:t-iss}}
    \subfloat[\(\tau\)-style score]
    {\includegraphics[width=0.23\textwidth,height=0.2\textwidth]{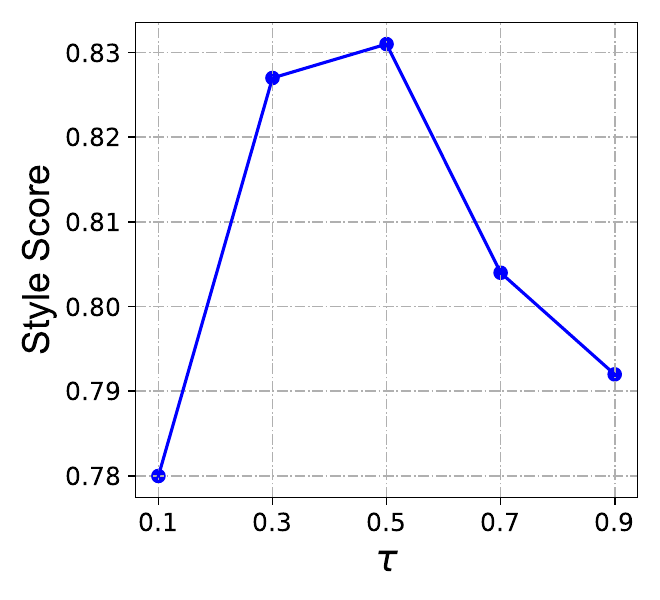}
    \label{fig:t-ss}}
    \subfloat[\(\tau\)-semantic score]
    {\includegraphics[width=0.23\textwidth,height=0.2\textwidth]{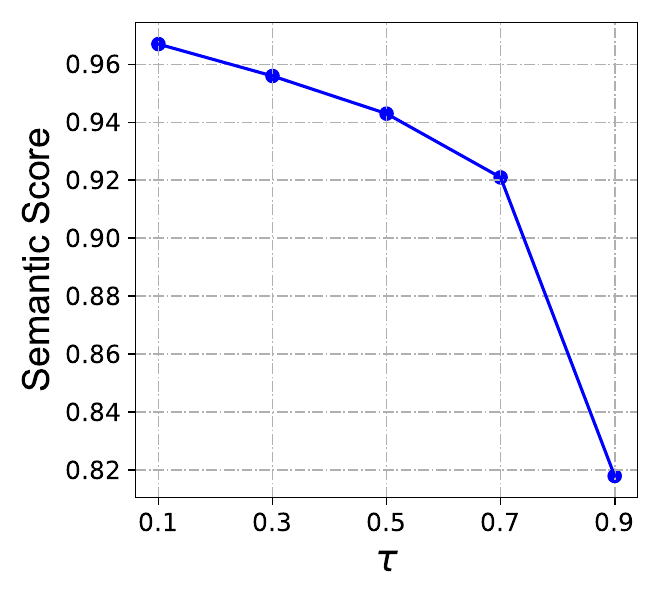}
    \label{fig:t-sts}}
    
    \subfloat[\(\lambda\)-issues]
    {\includegraphics[width=0.23\textwidth,height=0.2\textwidth]{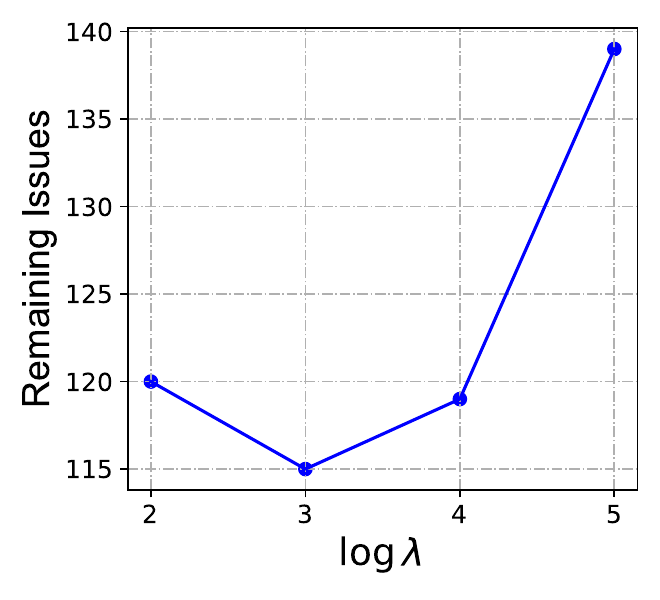}
    \label{fig:l-iss}}
    \subfloat[\(\lambda\)-style score]
    {\includegraphics[width=0.23\textwidth,height=0.2\textwidth]{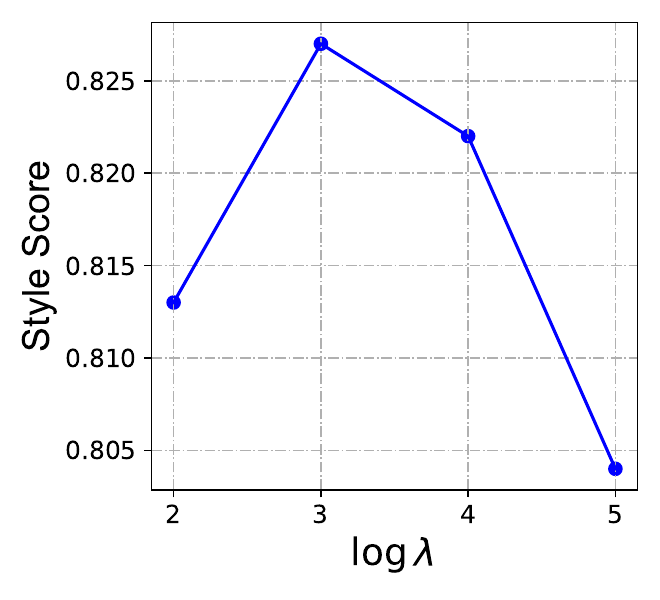}
    \label{fig:l-ss}}
    \subfloat[\(\lambda\)-semantic score]
    {\includegraphics[width=0.23\textwidth,height=0.2\textwidth]{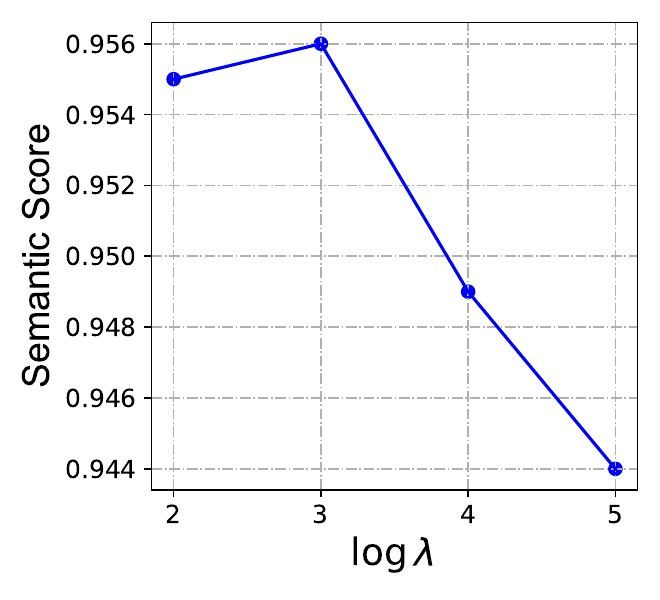}
    \label{fig:l-sts}}
    }
  \caption{Effect of \(\tau\) and \(\lambda\) on the average performance of \sys's repairing process.}
  \label{fig:tl}
\end{figure*}

\noindent
{\bf Experiment Design}:
\sys leverages three hyperparameters: detection threshold \(h\), inference temperature \(\tau\), and guidance strength \(\lambda\), to find and repair stylistic consistent issues.
The detection threshold \(h\) determines the point at which a sentence's style score is categorized as a stylistic inconsistent issue, with a lower \(h\) indicating a greater tolerance for style inconsistencies.
The inference temperature \(\tau\) represents the maximum lexical deviation allowed from the initial translation when generating the revised sentence, with a higher \(\tau\) granting greater freedom to modify the initial translation.
The parameter \(\lambda\) denotes the strength of user-supplied style guidance samples for generating revised sentences.

We conduct experiments to investigate and understand how different values of these configurable hyperparameters affect the performance of \sys in finding and repairing stylistic consistent issues.
Specifically, we evaluate \sys's performance using output from Google Translate, testing a range of \(h\) values from 0.3 to 0.8, \(\tau\) values from 0.1 to 0.9, and \(\lambda\) values from 1e2 to 1e5.

To assess the impact of \(h\) on the detection of style problems, we analyze changes in precision and false positive rate of the detector as \(h\) varies, using manual labeled samples as detailed in~\autoref{sec:rq1}.
For the style applicator, we evaluate the effects of varying \(\tau\) and \(\lambda\) on repair effectiveness, measuring changes in the number of repaired issues, overall style scores, and semantic textual similarity~(STS) values post-repair.

\noindent
{\bf Results}:
\autoref{fig:h} illustrates the impact of parameter related to the style detector on its performance, whereas \autoref{fig:tl} demonstrates the influence of parameters related to the style applicator.

\textit{Impact of \(h\):}
The parameter \(h\) influences the sensitivity of style detector in identifying stylistic inconsistencies.
As shown in~\autoref{fig:h-fpr}, FPR increases from 3\% to 18.3\% as \(h\) increases from 0.3 to 0.8, indicating that higher \(h\) value leads to a more radical detection of style issues.
Concurrently, precision increases from 80.6\% to a peak of 90.5\% at \(h=0.5\), then slightly increases to 92.2\% at \(h=0.8\).
The results indicate that both metrics increase as \(h\) rises, initially grows more slowly and then accelerates, while the precision increases rapidly at first and then plateaus.
This pattern suggests an optimal balance point at \(h=0.5\), where precision is nearly maximized while the false positive rate is reasonably low.

\textit{Impact of \(\tau\):}
The parameter \(\tau\) plays a crucial role in the repair of stylistic inconsistent issues.
\autoref{fig:tl}(a), (b) and (c) shows the effect of \(\tau\).
As \(\tau\) increases, the number of remaining issues after repair initially decreases, reaching an optimal value at \(\tau=0.3\), and then increases. 
Concurrently, the overall style score of the revised output follows a similar trend, achieving optimal performance at \(\tau=0.5\).
The semantic score, however, consistently decreases with increasing \(\tau\), with a more rapid decline observed at higher \(\tau\) values.
Considering the trade-off between style and semantic score, \sys selects 0.3 as the default value of \(\tau\).

\textit{Impact of \(\lambda\):}
The parameter \(\lambda\) affects the weight given to style preservation during the repair process.
\autoref{fig:tl}(d), (e) and (f) shows the effect of \(\lambda\).
The figure demonstrates that as \(\lambda\) increases, the number of residual issues after repair initially decreases and then increases. 
Similarly, both the overall style score and semantic score of the revised output follow an increasing trend initially, reaching an optimal point at \(\lambda=1000\), before declining. 
Consequently, \(\lambda=1000\) is selected as the default value for optimal performance.

\noindent
{\bf Analysis}:
From \autoref{fig:h} and \autoref{fig:tl}, we can observe that the configurable parameters \(h\), \(\tau\), and \(\lambda\) have a significant impact on the performance of \sys in detecting and repairing stylistic inconsistencies. 
In terms of detection accuracy, increasing 
h initially improves precision while maintaining a reasonable false positive rate, suggesting an optimal balance at \(h=0.5\). 
For repair performance, the parameter \(\tau\) shows that allowing moderate lexical deviations (\(\tau=0.3\)) optimizes the number of corrected stylistic issues, while a higher \(\tau\) value can detrimentally affect semantic integrity. 
The guidance strength parameter \(\lambda\) demonstrates that moderate user-supplied guidance (\(\lambda=1000\)) enhances both stylistic and semantic scores, with performance declining at higher values.
Consequently, to achieve optimal detection and repair, we set \(h=0.5\), \(\tau=0.3\) and \(\lambda=1000\) as default values in \sys.

\noindent
{\bf Summarization}:
We have proved the advancement of \sys through the above experimental evaluations.
Overall, \sys is capable of detecting over 80\% of stylistic inconsistencies in translations and successfully enhances approximately 83\% of these inconsistent outputs.
The additional computational expense of \sys remains relatively modest, averaging no more than 6 seconds, which makes it feasible for integration into a wide range of commercial translation systems.

\section{Qualitative examples}\label{sec:examples}

\begin{figure*}[h]
    \centering
    \scalebox{0.9}{
    \includegraphics[trim={1.8cm 9.4cm 1.8cm 2.3cm},clip,width=\textwidth]{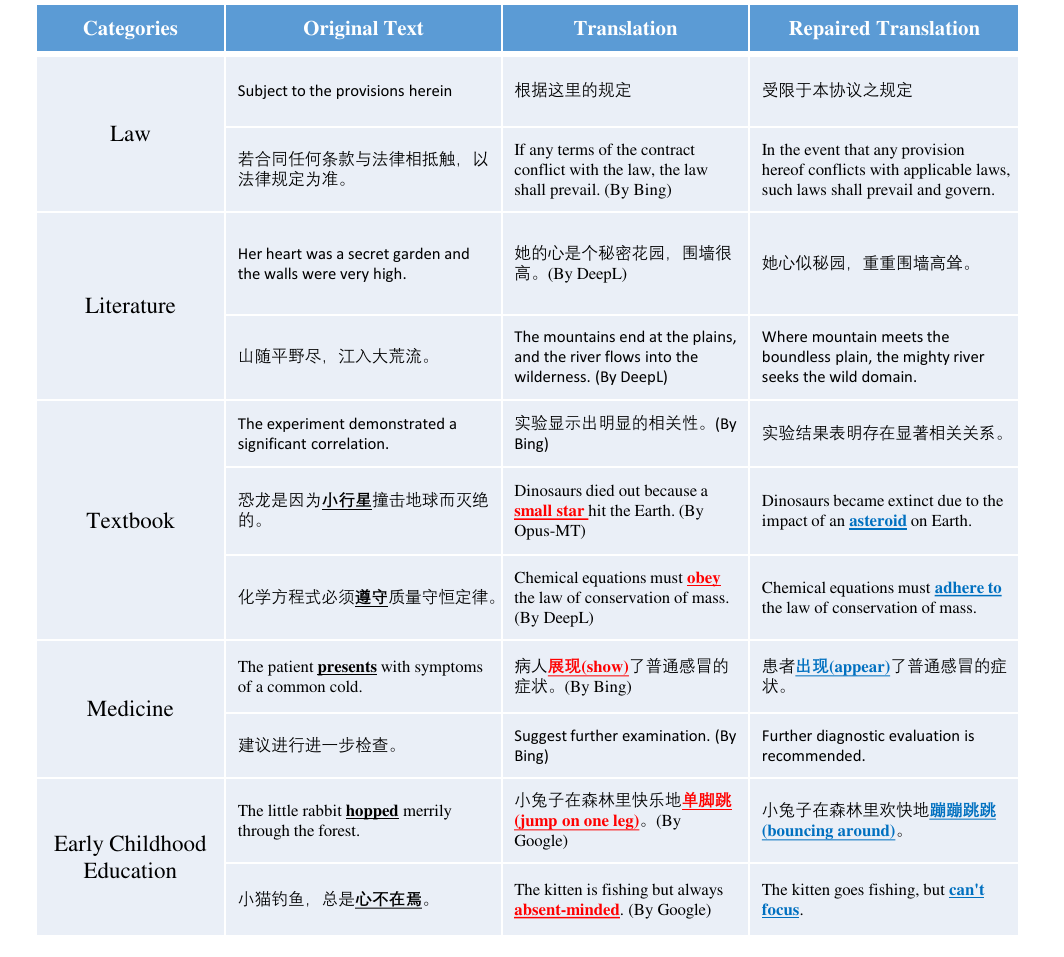}
    }
    \caption{Example of stylistic inconsistent issues and repaired translation generated by \sys.}
    \label{fig:case}
\end{figure*}

To illustrate the effectiveness of \sys in preserving domain-specific styles, we present a collection of example translations in \autoref{fig:case}. 
These examples span our five domains (legal, literary, scientific writing, medical, and educational content) and demonstrate both Chinese-to-English and English-to-Chinese translations. 
Each row shows an original text, its direct translation from a commercial system (Google Translate, Youdao Translate, Baidu Translate, or Opus-MT), and \sys's style-refined version. 
For instance, in legal texts, \sys transforms casual expressions like ``\begin{CJK*}{UTF8}{gbsn}根据这里的规定\end{CJK*}'' into proper legal language ``\begin{CJK*}{UTF8}{gbsn}受限于本协议之规定\end{CJK*}'', maintaining formal register. 
In literary translation, it preserves poetic elements, transforming literal translations like "The mountains end at the plains" into more literary renderings like "Where mountain meets the boundless plain". 
The examples highlight how \sys preserves domain-appropriate terminology and conventions while maintaining semantic accuracy. 
Bold text indicates specific stylistic elements that were improved in the repair process.

\end{document}